\documentclass[journal]{IEEEtai}

\usepackage[colorlinks,urlcolor=blue,linkcolor=blue,citecolor=blue]{hyperref}

\usepackage{color,array,amsmath,bm}
\usepackage{algorithm}
\usepackage{algorithmic}
\usepackage{graphicx}
\usepackage{overpic}
\usepackage{soul} 
\usepackage{graphicx}
\usepackage{multirow}
\usepackage{booktabs}
\usepackage[a4paper, total={7in, 10in}]{geometry}
\usepackage[font=footnotesize]{caption}
\makeatletter
\newcommand{\removelatexerror}{\let\@latex@error\@gobble}
\makeatother

\begin{document}

\title{ Brain-inspired Multi-scale Evolutionary Neural Architecture Search for Deep Spiking Neural Networks} 
\author{Wenxuan Pan, Feifei Zhao, Guobin Shen, Bing Han,Yi Zeng

\thanks{Wenxuan Pan is with the Brain-inspired Cognitive Intelligence Lab, Institute of Automation, Chinese Academy of Sciences, Beijing 100190, China, and School of Artificial Intelligence, University of Chinese Academy of Sciences, Beijing 100049, China.} 
\thanks{Feifei Zhao is with the Brain-inspired Cognitive Intelligence Lab, Institute of Automation, Chinese Academy of Sciences, Beijing 100190, China.} 
\thanks{Guobin Shen is with the Brain-inspired Cognitive Intelligence Lab, Institute of Automation, Chinese Academy of Sciences, Beijing 100190, China, and School of Future Technology, University of Chinese Academy of Sciences, Beijing 100049, China.} 
\thanks{Bing Han is with the Brain-inspired Cognitive Intelligence Lab, Institute of Automation, Chinese Academy of Sciences, Beijing 100190, China, and School of Artificial Intelligence, University of Chinese Academy of Sciences, Beijing 100049, China.}
\thanks{Yi Zeng is with the Brain-inspired Cognitive Intelligence Lab, Institute of Automation, Chinese Academy of Sciences, Beijing 100190, China, and University of Chinese Academy of Sciences, Beijing 100049, China, and Center for Excellence in Brain Science and Intelligence Technology, Chinese Academy of Sciences, Shanghai 200031, China.}

\thanks{The first and the second authors contributed equally to this work, and serve as co-first authors.}
\thanks{The corresponding author is Yi Zeng (e-mail: yi.zeng@ia.ac.cn).}}

\maketitle

\begin{abstract}
Spiking Neural Networks (SNNs) have received considerable attention not only for their superiority in energy efficiency with discrete signal processing but also for their natural suitability to integrate multi-scale biological plasticity. However, most SNNs directly adopt the structure of the well-established Deep Neural Networks (DNNs), and rarely automatically design Neural Architecture Search (NAS) for SNNs.
The neural motifs topology, modular regional structure and global cross-brain region connection of the human brain are the product of natural evolution and can serve as a perfect reference for designing brain-inspired SNN architecture. In this paper, we propose a Multi-Scale Evolutionary Neural Architecture Search (MSE-NAS) for SNN, simultaneously considering micro-, meso- and macro-scale brain topologies as the evolutionary search space. MSE-NAS evolves individual neuron operation, self-organized integration of multiple circuit motifs, and global connectivity across motifs through a Brain-inspired Indirect Evaluation (BIE) Function. 
This training-free fitness function could greatly reduce computational consumption and NAS's time, and its task-independent property enables the searched SNNs to exhibit excellent transferability on multiple datasets. 
Furthermore, MSE-NAS show robustness against the training method and noise.
Extensive experiments demonstrate that the proposed algorithm achieves state-of-the-art (SOTA) performance with shorter simulation steps on static datasets (CIFAR10, CIFAR100) and neuromorphic datasets (CIFAR10-DVS and DVS128-Gesture). 
The thorough analysis also illustrates the significant performance improvement and consistent bio-interpretability deriving from the topological evolution at different scales and the BIE function.

\end{abstract}


\section{Introduction}
Through millions of years of evolution, the brain has evolved specific neural circuits and modular regional structures, enabling it with the ability to integrate multi-scale information processing ranging from microscopic neuronal interactions to macroscopic collaborations between various brain regions~\cite{luo2021architectures,pan2020activity,isaacson2011inhibition}. 

As an event-driven, low-power third-generation neural network, Spiking Neural Network (SNN) can simulate the discrete communication among biological neurons and incorporate various biological plasticity learning rules and brain-inspired neural network architectures~\cite{maass1997networks}. Therefore, SNNs have gained considerable attention in recent years.

Most of the research on SNNs has focused on the training and optimization of SNNs, ranging from unsupervised training of shallow SNNs~\cite{diehl2015fast} to indirect optimization of SNNs based on artificial neural networks (ANNs) conversion~\cite{diehl2015fast}, to the recent direct training of SNNs based on surrogate gradient~\cite{lee2016training,lee2020enabling,gu2019stca,wu2021training,wu2018spatio}. 
As can be seen from many successful hand-designed network architectures~\cite{he2016deep,simonyan2014very,vaswani2017attention}, different structures can bring different functions to the model. However, the efficiency of some attempts to transfer the structure of deep neural networks (DNNs) to spiking neural networks (e.g. transformer~\cite{vaswani2017attention}, VGG-Net~\cite{sengupta2019going}, ResNet~\cite{he2016deep}, inception~\cite{szegedy2015going}) still lags behind that of ANNs. 
Different from ANNs, SNNs accumulate information through potential accumulation and transmit information through spike emission, which leads to the fact that the architecture adopted for ANN may not necessarily match the working mechanism of SNN.
Another recurrent randomly connected SNN architecture is called reservoir~\cite{maass2002real} with a non-hierarchical topology consistent with the connectome found in the brain~\cite{suarez2021learning}. Reservoirs-based SNN architectures are difficult to be optimized and thus couldn't achieve comparable performance to DNNs. SNNs require customized architectures to guarantee performance while exhibiting high biological interpretability.

Rather than time-consuming, artificially designed network architectures, Neural Architecture Search (NAS) can automatically design appropriate neural architectures for specific tasks and thus has been broadly applied in DNNs.
Existing NAS algorithms include but are not limited to gradient-based algorithms~\cite{pham2018efficient,liu2018darts,zela2019understanding}, reinforcement learning-based algorithms~\cite{zoph2016neural,zoph2018learning,baker2016designing} , and 
evolution-based algorithms~\cite{real2017large,xie2017genetic,wen2021two}.
Among them,  Evolutionary Neural Architecture Search (ENAS) ~\cite{floreano2008neuroevolution,liu2021survey,zhu2021real,wen2021two,sun2019evolving,zhang2022evolutionary,xie2022benchenas} enables experts to design without prior knowledge, without building supernets in advance or high computational costs but drives large-scale iterative updates of populations through evolutionary paradigms to obtain the optimal global solution. Evolutionary searching for a brain-inspired efficient SNN architecture is of great significance for improving the information processing capacity of SNN and bridging the gap between artificial and natural evolution~\cite{stanley2019designing,stockl2021structure}.

To the best of our knowledge, there has been little work on optimizing SNNs with ENAS. 
\cite{na2022autosnn} follows the experience of VGG-Net and ResNet, searching for the stacking rule of spiking convolution block and spiking residual block to generate SNN architectures. \cite{kim2022neural} looks for cell-based SNN architectures, but focusing on only one backward connection makes the search space less rich and flexible. 
Existing works on evolving SNN architectures currently lack in-depth inspiration from the multi-scale brain neural topology, and the performance needs to be improved. 
This paper takes inspiration from the evolutionary mechanisms of brain multi-scale neural circuits and aims to rapidly and automatically evolve brain-inspired architectures for high-performance SNNs.

The human brain nervous system is integrated with multi-scale topology, ranging from information processing in individual neurons (micro-scale) to the topology of neural microcircuits among neurons (mesoscopic scale), to long-term connectivity across brain regions (macro-scale)~\cite{luo2021architectures}. 
Drawing inspiration from the natural evolutionary mechanism of brain multi-scale neural microcircuits and multi-brain regions coordinated neural architectures, this paper introduces a Multi-Scale Evolutionary Neural Architecture Search (MSE-NAS) algorithm for brain-inspired spiking neural networks. 
Unlike previous methods that search for basic operational components or DNN-based block units, we evolve self-organised coordination between neural motifs topology and long-term global connections across motifs. 
In addition, to reduce the computational cost of evaluating individual fitness during evolution, we design a brain-inspired indirect evaluation method (instead of the costly accuracy) inspired by the stability of neural activity in the brain~\cite{luo2021architectures,kriegeskorte2008representational,maass2004methods}, greatly reducing computational cost and helping to search for diverse and efficient network structures. To the best of our knowledge, this is the first work to draw on the neural architecture obtained from the natural evolution of the brain, simultaneously evolving the brain's microscopic, mesoscopic, and macroscopic topologies for SNN. 

Overall, the contributions made in this paper can be summarized as the following points:
\begin{itemize}

\item[1)] This paper evolves the organization of multiple common neural circuit motifs in the biological brain that are composed of varied combinations of excitatory and inhibitory connections. 
We demonstrate that the coordinated combination between different neural microcircuits enables the network architecture to evolve into a complex spiking neural network with excitation-inhibition interleaving, endowing the SNNs with a powerful learning and memory capacity.

\item[2)] This paper adopts a genetic algorithm to simultaneously evolve microscopic, mesoscopic, and macroscopic topologies for SNN. Microscopically, MSE-NAS searches operations for each neuron in SNN. Mesoscopically, MSE-NAS evolves coordinated combinations of multiple common neural circuit motifs to generate brain-inspired motif topology-based neural architectures. Macroscopically, the number of motifs and the cross-layer global connections between motifs are regarded as evolvable objects, enabling the model flexibly change the volume to adaptively meet the needs of the task.

\item[3)] We propose a rapid brain-inspired evaluation method (instead of costly accuracy), namely BIE to evaluate the stability of individual based on the neural activity. This training-free evaluation method dramatically reduces the computational cost and saves the time of architecture search, which is of great significance for evolutionary NAS that require extensive computational evaluation.

\item[4)] Extensive comparative experiments on multiple static (CIFAR10, CIFAR100) and neuromorphic (CIFAR10-DVS and DVS128-Gesture) datasets demonstrate the superior state-of-the-art (SOTA) performance of our proposed SNN-based MSE-NAS algorithm. The ablation analysis further demonstrates the effectiveness of motif-based self-organising architectural evolution and cross-layer connectivity evolution.
Furthermore, the proposed model shows significant transferability, robustness, and energy advantages, exhibiting strong generalization performance.

\end{itemize}

This article is organised as follows. Section~\ref{rel} introduces the existing related work. Section~\ref{pro} explains the proposed multi-scale evolutionary neural architecture search method for SNNs in detail, and then Section~\ref{exp} shows the experimental results to prove the validity of the model and various conclusions. Finally, the full manuscript is discussed and summarised in Section~\ref{con}.

\section{Related work} \label{rel}

In recent years, NAS has attracted great interest due to its automatic design of high-performance and robust network architectures instead of costly handcraft DNN architectures. ~\cite{zoph2016neural} and \cite{baker2016designing} use the reinforcement learning method to search for the best neural architecture, considered to be the pioneering works of NAS. Subsequently, many researchers have applied NAS to various fields~\cite{liu2019auto,zhang2019customizable,ghiasi2019fpn}, and how to balance NAS efficiency and cost has also aroused the interest of researchers~\cite{li2019partial, fedorov2019sparse}.

\subsection{Evolutionary Neural Architecture Search}

Existing NAS algorithms can be roughly divided into gradient-based~\cite{pham2018efficient,liu2018darts,zela2019understanding}, reinforcement learning-based~\cite{zoph2016neural,zoph2018learning,baker2016designing} and evolutionary computing-based algorithms~\cite{real2017large,xie2017genetic,wistuba2019deep,elsken2017simple,chen2019progressive}. 
Gradient-based algorithms are very costly to implement because they often find ill-conditioned architectures and need to build supernets in advance, while reinforcement learning-based algorithms are considered pioneers in the field of NAS, but their high computational resource costs (hundreds of GPU days or more) often discourage ordinary researchers. 
Inspired by the natural selection law of survival of the fittest, Evolutionary Neural Architecture Search (ENAS) can avoid these problems and obtain the global optimal solution more easily. \cite{real2017large} introduces evolution into NAS for the first time, and also achieved the most advanced results at that time. \cite{real2017large,xie2017genetic} encode the neural network architecture into a string and treats it as a chromosome in the population. Then, genetic operators such as crossover mutation are applied to the population, and the population is iteratively updated according to the fitness of each individual, and the survival of the fittest is achieved. 

\subsection{Search Space}

Due to the wide range of network architectures that can be encoded, research is gradually developing towards more efficient and modular architectures. 
Common search spaces can be divided into cell-based~\cite{zhong2018practical,liu2017hierarchical,dong2018dpp} and block-based~\cite{cui2019fast,wu2019fbnet} search spaces. 
In block-based search spaces, the structure of blocks can be diverse. Through the experience of learning successful DNN architecture, ResBlock~\cite{he2016deep}, DenseBlock~\cite{huang2017densely}, InceptionBlock~\cite{szegedy2015going} and other blocks~\cite{howard2019searching} with good performance are designed and organized, which reduces the number of parameters required for searching and improves network efficiency~\cite{howard2019searching,chen2019auto,song2020efficient}. 
\cite{cui2019fast} proves that stacking diverse building blocks is beneficial to improving network performance, and its computational cost is low. By summarising the rules of excellent artificially designed architectures, NASNet~\cite{zoph2018learning} finds that repeated stacking of multiple small cells can improve model performance and reduce the amount of calculation. 

In summary, NAS research on DNNs exploits efficient modular structures to help reduce the parameter space for searching~\cite{liu2021survey}. However, the computing unit or block is still a manually defined DNN architecture. The human brain has evolved over hundreds of millions of years to form naturally with modular topologies. Developing NAS by directly drawing on the multi-scale neural topology evolved from the brain is a promising approach to exploring high-performance and low-energy brain-inspired SNNs.

\subsection{Neural Architecture Search for Spiking Neural Networks}
SNNs exhibit a high biological plausibility and computational efficiency that stems from information processing with binary spikes.
The spiking neurons receive signals from presynaptic neurons, integrate them to form postsynaptic potentials, and deliver spikes when the membrane potential exceeds a threshold~\cite{maass1997networks}. 
Except that individual neurons can perform complex nonlinear information processing, the synaptic connectivity between neurons enables neurons to form functional neural microcircuits and even more complex brain nervous systems.
SNNs serve as more brain-inspired third-generation artificial neural networks and have the potential to integrate biological information processing, connectivity structures, and cognitive functions to achieve more powerful and general artificial intelligence.

However, there is still a lack of research on optimising brain-inspired SNN architectures. Most of the SNNs' architectures follow the successful structure of DNNs, such as residual structures ~\cite{he2016deep}, inception module~\cite{szegedy2015going}, transformer~\cite{vaswani2017attention}. \cite{na2022autosnn} adopt the experience of VGG-Net and ResNet, searching for the stacking rule of spiking convolution block and spiking residual block to generate SNNs for different tasks. It customises the search space from the perspective of reducing energy consumption and searches for a more suitable architecture for SNN through evolutionary algorithms, but the performance is still weaker than comparative ANN models. \cite{kim2022neural} looks for cell-based SNN architectures, but focusing on only one backward connection makes the search less comprehensive. Current efforts in evolving SNN architectures fall short of drawing profound inspiration from the multi-scale brain neural topology, leaving room for performance enhancements.

\subsection{Fitness Evaluation}
At the same time, ENAS requires evaluating each fitness to optimise the choice of architecture. Most methods regard the model's classification accuracy on the dataset as an individual's fitness, which is extremely time-consuming. Therefore, some performance predictors use the acquired knowledge to make reasonable predictions on the performance of individuals, which greatly saves the time required for evaluation.~\cite{sun2019surrogate} propose an end-to-end performance predictor based on random forest, accelerating the evolution process of DNNs.~\cite{abdelfattah2021zero} propose a zero-cost proxy method that can predict model performance without training, and experiments have shown that it can enhance or accelerate many existing NAS algorithms. However, these methods are based on DNNs, and efficient indirect evaluation methods for SNNs architectures are still scarce. 

Although NAS algorithms have achieved remarkable success in DNNs, there is still a lack of exploration of SNNs. The only NAS algorithms for SNNs also depend on the handcraft DNNs architecture. In this paper, we take inspiration from the multi-scale neural topology in the brain and integrate natural evolutionary principles into the NAS for SNN. Besides, to reduce computational consumption, a brain-inspired indirect evaluation method without training is designed to endow ENAS of SNN with the capability of efficient architecture search, high performance, transferability on multiple datasets and robustness to noise.
\begin{figure*}[htp]
\centering
\includegraphics[width=16cm]{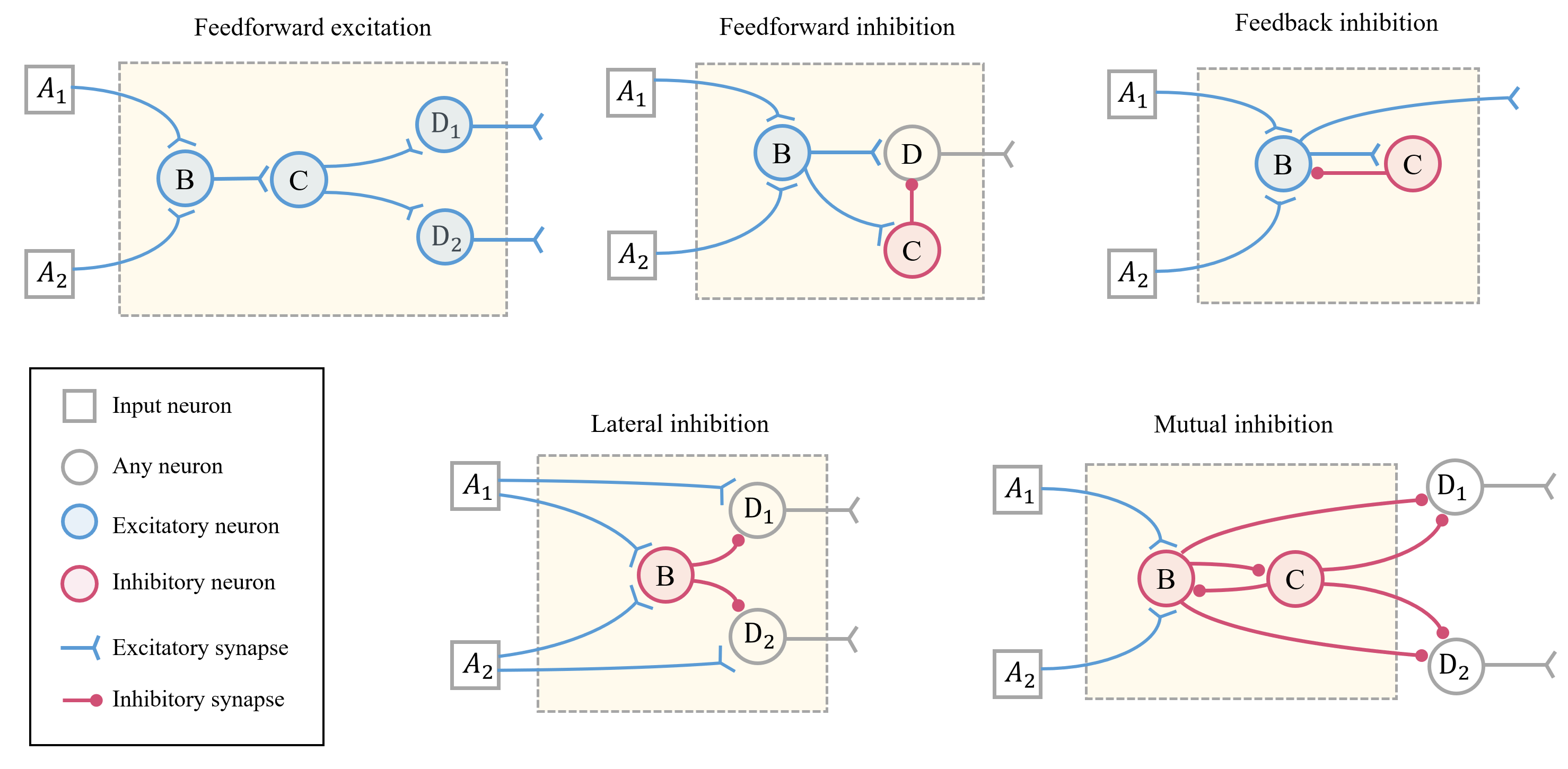}
\caption{Basis of architecture evolution: common neural microcircuits. Blue represents excitatory neurons and synapses, and red represents inhibitory neurons and synapses. }
\label{fige1}
\end{figure*}

\section{ Multi-scale Evolutionary Neural Architecture Search for SNNs}\label{pro}
Over hundreds of millions of years of evolution, neural motifs with specific functions are established through synaptic connections and provide a physical basis for the cooperative computation between neurons and the complex information processing mechanism of the nervous system~\cite{luo2021architectures}. 
From microscopic neurons to macroscopic neural motifs to global cross-brain region connections, multiple scales of topological connectivity in the nervous system can be clearly observed, as shown in Fig.~\ref{fig2}a.

We adopt the genetic algorithm to optimize SNN architectures from multiple scales and decompose the search space into micro, meso, and macro: taking spiking neurons as the foundational computing unit, constructing functional neural motifs and exploring the cross-layer global architecture design of SNN through the connection mode between motifs.

Evolving a neural network architecture takes a lot of time. The traditional way of directly measuring the classification accuracy of a network under a specific dataset is time-consuming. Moreover, for its biological plausibility, SNN needs some time steps to accumulate spatiotemporal information, which makes the process of finding the optimal architecture longer. A suitable evaluation method for evolutionary spiking neural architecture search not only can reduce the high computational cost of extensive network architecture search but also help to select more efficient and diverse SNN architectures. 
In order to save time costs, 
we design a brain-inspired indirect evaluation method without training to assess individual fitness.

This section first introduces the basic components of the SNN architecture involved in this paper and then detailed describes the two highlights of our proposed MSE-NAS algorithm: multi-scale evolutionary coding space and indirect fitness evaluation method.
\subsection{Spiking Neural Network Foundations}

\subsubsection{Leaky Integrate-and-Fire Neuron}
In the brain's nervous system, neurons transmit information by firing spikes. A commonly used neuron model is called Leaky Integrate-and-Fire (LIF) neuron~\cite{lapicque1907recherches}, and its membrane potential is updated with the input formula as \eqref{eq1}:

\begin{align}
\label{eq1}
\tau_{\mathrm{m}} \frac{d V_{\mathrm{m}}(t)}{d t}&=I(t)-V_{\mathrm{m}}(t)
\end{align}%
where $I(t)$ represents the external input received by the neuron at time $t$, $V_m(t)$ represents the membrane potential of the neuron at time $t$, and $\tau=2.0$ is the membrane potential time constant. The spiking neuron integrates the input of the upstream neuron and its own activity, calculates the current membrane potential through the above formula \eqref{eq1}, and decides whether to send a spike to the postsynaptic neuron according to whether the membrane potential exceeds $V_{th}=0.5$. The LIF neuron model in this work is implemented on Brain-inspired Cognitive Intelligence Engine (BrainCog) framework~\cite{zeng2022braincog}.

\subsubsection{Neural Microcircuits}

\begin{figure*}[htp]
\centering
\includegraphics[width=18cm]{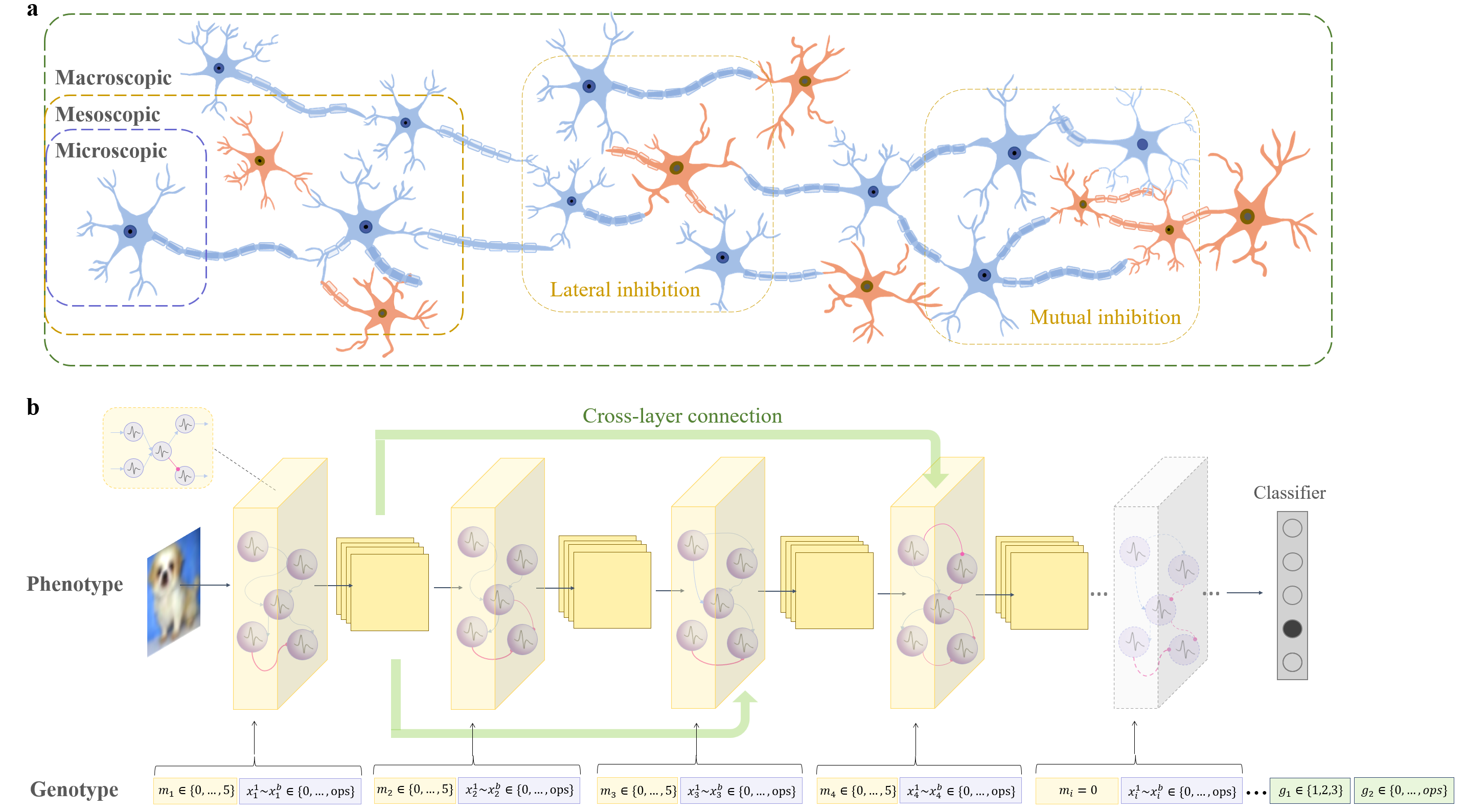}

\caption{Multi-scale Encoding Space. \textbf{a.} Multi-scale topological connectivity in the brain, from the microscopic neuron scale (purple), the mesoscopic neural microcircuit scale (yellow) to the global macroscopic connectivity (green). Blue neurons are excitatory neurons and orange neurons are inhibitory neurons. \textbf{b.} Genotypes and phenotypes of evolved NAS of SNN inspired by brain multi-scale architecture.}
\label{fig2}
\end{figure*}

In SNNs, spiking signals are transmitted through varied neural circuits formed between neurons, and the diverse excitatory-inhibitory connections that make up the neural circuits endow the network with distinct functionalities.
Some neural microcircuits may have originated a long time ago, and many functional circuit motifs have been preserved in the human brain through evolution. Among them, the five most common motifs are summarized in~\cite{luo2021architectures}, as shown in Fig.~\ref{fige1}:

(1) Feedforward Excitation (FE). Excitatory neuron $B$ receives multiple excitatory signals (incoming from $A$ in Fig.~\ref{fige1}) is called convergent excitation, which enables neurons to selectively respond to simulations of diverse modalities and can integrate multiple simultaneous inputs of various signals~\cite{meredith2002neuronal} in the nervous system.
$C$ receives input from $B$ through branched axons and outputs to multiple postsynaptic neurons, called divergent excitation which helps to enhance the signal-to-noise ratio when the input signal carries noise~\cite{luo2021architectures}. 

(2) Feedforward Inhibition (FI). The excitatory signal is transmitted to the inhibitory neuron $C$, causing the postsynaptic neuron $B$ to receive the inhibitory signal. Inhibitory signalling is essential for locally modulating the amplitude and duration of excitatory long-range signals~\cite{sherrington1925remarks,coombs1955inhibitory}. Synaptic inhibition often occurs simultaneously with excitation. The coordinated adjustment of the two can prevent excessive excitation or inhibition, maintaining the dynamic balance of the neural network~\cite{isaacson2011inhibition}.

(3) Feedback Inhibition (FbI). The excitatory neuron $B$ is activated, and the inhibitory neuron $C$, as the post-synaptic neuron of $B$, sends an inhibitory signal back to $B$ after receiving the excitatory signal coming from $B$. Like feedforward inhibition, feedback inhibition can also modulate excitatory signals, enhancing the signal-to-noise ratio in information processing~\cite{luo2021architectures}.

(4) Lateral Inhibition (LI). Inhibitory neuron $C$ receives stimuli from multiple pathways and transmits inhibitory signals to multiple postsynaptic neurons. Lateral inhibition prevents the spread of incoming excitatory potentials, accentuating contrast by amplifying their differences, thereby sharpening the perception of receptors~\cite{matlin1992sensation,orchard1991neural}. This phenomenon is found in vision, hearing and smell~\cite{laurent2001odor,mountcastle1957modality}.

(5) Mutual Inhibition (MI). Neurons $B$ and $C$ inhibit each other, and signals may also be transmitted to other neurons. Mutual inhibitory circuits underlie rhythmic neuronal activity, manifested in locomotion~\cite{grillner2006biological} and sleep state switching~\cite{saper2010sleep}.

In this paper, we use LIF neurons to model the spiking neurons within different motifs and combine excitatory and inhibitory connections to form circuit motifs with different structures and functions. $A$ represents feature maps transmitted to this layer by the previous layer (or multiple layers if there are cross-layer connections).

\subsection{Evolutionary Neural Architecture Search}

\subsubsection{Multi-scale Encoding Space}
Different SNN architectures need to be encoded as genotypes to implement evolutionary spiking neural network architecture search. Fig.~\ref{fig2}a shows that the multi-scale encoding proposed in this paper is inspired by the microscopic (neuron), mesoscopic (neural microcircuit) and macroscopic (cross-brain region connection) computational scales that exist in the nervous system.

Fig.~\ref{fig2}b depicts the mapping relationship between genotype and phenotype. In general, the genotype of an SNN architecture can be divided into three parts: $[X,M,G]$, which represent the three scales of evolution respectively. Assuming the model has at most $l$ layers, the genotype (total length $(l*(b+1)+2)$) is converted to phenotype using a fixed-length encoding method, where each layer is encoded in $b$ genes. $b$ is a fixed length, set to 20 here to fully represent all microscopic operations in any motif, including operation type and the type of neuron (excitatory or inhibitory). The first $l*(b+1)$ genes indicate the micro and meso scales and the last two genes indicate the macro scale. 

Take layer $i$ as an example to illustrate the evolution scheme of micro and meso levels. It is encoded as $(m_i,x_i^1,...x_i^{b})$. At the microscopic level, the evolvable operations for LIF neurons include 3x3 convolution and 5x5 convolution, encoded by $x_i^1$ to $x_i^b$ genes. As a mesoscopic scale, a neural microcircuit is defined as an SNN layer with multiple LIF neurons inside. The first gene $m_i \in \{1,2,3,4,5\}$ indicates which microcircuit forms the layer $i$. Especially, $m_i=0$ means that the layer is empty. Five neural microcircuits that can be evolved are shown in Fig.~\ref{fige1}.

Macroscopically, there is an evolvable layer depth and global connection mode - each layer can selectively receive the output of the previous layer, the first two layers or the first three layers as input (as shown in the cross-layer connection ~\ref{fig2}b), represented by the last two genes $g_1,g_2$ of the genome. $g_1=1$ indicates that the entire network is an ordinary forward structure, and each layer receives the output of the previous motif as input. $g_1=2$ indicates accepting the output of the previous two motifs as input at the same time. $g_1=3$ indicates accepting the output of the previous three layers as input at the same time. The value of $g_2$ indicates the operation type of cross-layer connection, which can be 3x3 convolution or 5x5 convolution. It is worth mentioning that the value of $m_i$ also affects the macroevolution scale: the depth of the network is determined by the number of non-zero values in $m_1, m_2, m_3.., m_l$.

The genotype of MSE-NAS is encoded as:

\begin{align}
\label{geno}
{[X,M,G] = [(m_1,x_1^1,...x_1^b),...,(m_l,x_l^1,...x_l^{b}),g_1,g_2]}
\end{align}%
See the table \ref{ss} for the specific meaning of multi-scale space sampling values.

\begin{table}[htp]
\begin{tabular}{llll}
    \toprule
 Scale & Gene&Value & Description \\
    \midrule
 \multirow{2}{*}{Micro ($X$)} &\multirow{2}{*}{ $x_i^1 \sim x_i^b$ } &1 & 3x3 conv \\
&  & 2 &5x5 conv\\
\hline \multirow{6}{*}{Meso ($M$)}  &\multirow{6}{*}{$m_i$} 
& 0 &Empty \\
&  & 1 &Feedforward Excitation \\
&  & 2 &Feedforward Inhibition\\
&  & 3 &Feedback Inhibition\\
&  & 4 &Lateral Inhibition\\
&  & 5 &Mutual Inhibition\\

\hline \multirow{5}{*}{Macro ($G$)}  
& \multirow{3}{*}{$g_1$} &1 & Not across layers \\
&  & 2 &Connect across one layer\\
&  & 3 &Connect across two layer\\
\cmidrule{2-4}
& \multirow{2}{*}{$g_2$} &1 & 3x3 conv \\
&  & 2 &5x5 conv\\
\bottomrule

\end{tabular}
\caption{Detailed description of the meanings of different gene values in a genome.}\label{ss}
\end{table}

\subsubsection{Evolution Process}

During initialization, a population of SNN genotypes of size $n_{pop}$=50 (the rule is shown in Table~\ref{ss}) is randomly generated, and each genotype corresponds to an SNN architecture with max $l$ layers.

\begin{figure}[htp]
\centering
\includegraphics[width=8.5cm]{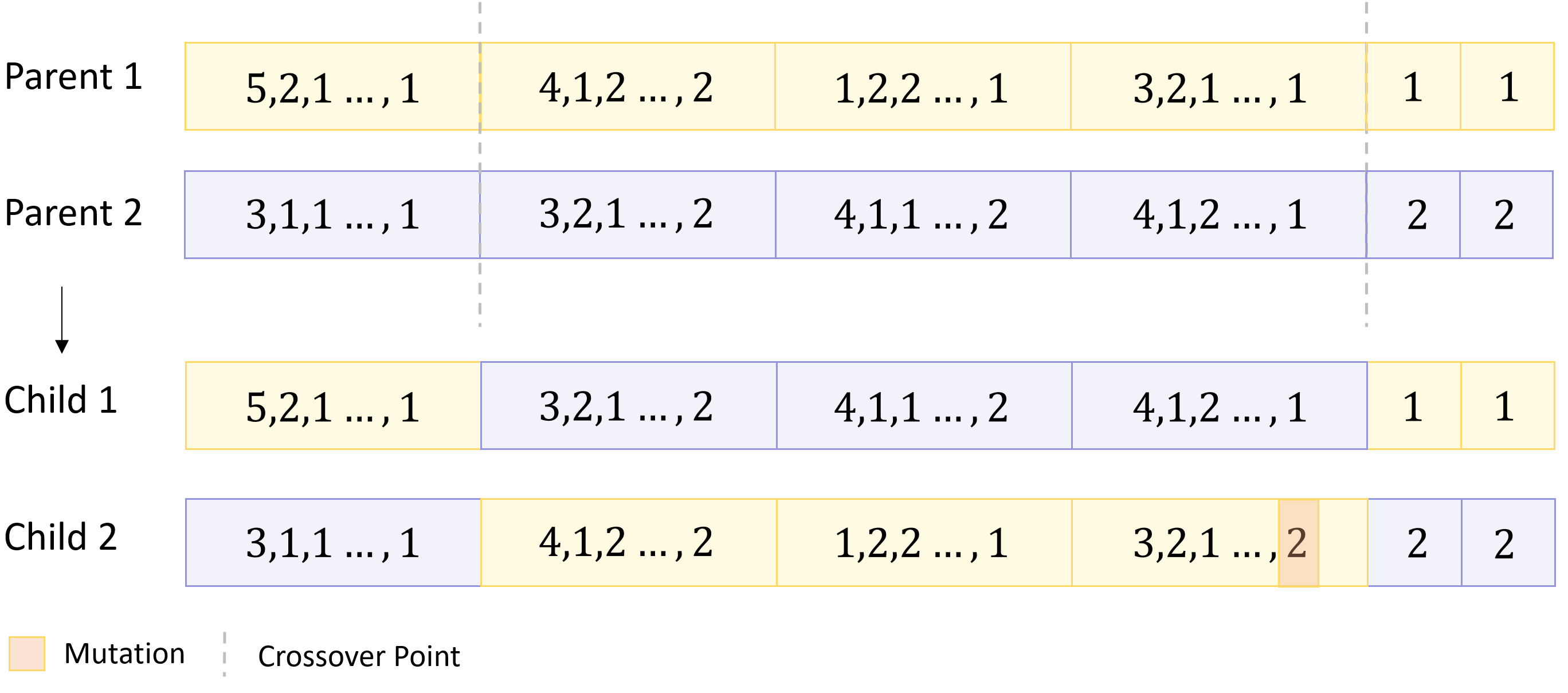}
\caption{Schematic diagram of crossover and mutation operators. }
\label{crossover}
\end{figure}

Since the microcosmic ($x_i^1$ to $x_i^b$) encodes the calculation inside the mesoscopic ($m_i$) motif, the two levels (micro and meso) are intimately intertwined, making cross-scale crossover meaningless.
Therefore, we restrict the genetic operator of two-point crossover to only existing meso-level and macro-level, which means that the crossover point can only be before or after a motif. 
The crossover probability is set to 0.3.
Bit-flip mutations applied with a probability of 0.02 act simultaneously on microscopic, mesoscopic, and macroscopic scales.
The way of crossover and mutation is shown in Fig.~\ref{crossover}.

We adopt the tournament selection operator to select part of the individuals as parents and generate the next generation of the population through two-point crossover and bit-flip mutation operators.
In particular, we propose a fitness evaluation method for SNN architectures as a criterion for generating the next generation of individuals. The evolution stops at the 80th generation, and the 10 individuals with the highest fitness scores are selected and trained for 20 epochs based on the surrogate gradient. The one with the highest classification efficiency is trained for 600 epochs to get the final performance. The whole process is shown in Algorithm~\ref{alg:evo}.

\begin{figure}[htp]

  \renewcommand{\algorithmicrequire}{\textbf{Input:}}
  \renewcommand{\algorithmicensure}{\textbf{Output:}}
  \removelatexerror
  \begin{algorithm}[H]

    \caption{The procedure of MSE-NAS algorithm.}
    \label{alg:evo}

    \begin{algorithmic}[1]
      \REQUIRE {$generations$, $n_{pop}$, $n_{offs}$, $l$, data}
      \ENSURE  {Best individual $Model$}
        \STATE {$pop$ = Initialize($n_{pop}$, $l$).}

      \WHILE {$g<generations$}

      \WHILE {$off.size<n_{offs}$}
      \STATE {$n_{select}$=$off.size$-$n_{offs}$}
      \STATE {$parents$ = TournamentSelection($pop$, $n_{select}$)}
      \STATE {$off$=PointCrossover($pop$, $parents$)}
      \STATE {$off$=PolynomialMutation($off$)}
      \ENDWHILE

      \STATE{$fit$=ComputeFitness($off$, data)}
      \STATE{$pop$=Select($pop$, $off$, $fit$, $n_{pop}$)}

      \ENDWHILE
      
      \STATE {$Genotypes$= Select($pop$,$fit$,$10$)}
      
        \STATE {$Phenotypes$ = Phenotype($Genotypes$)}
    \STATE {Ten models with the highest fitness scores $Phenotypes$ (with randomly initialized weights) are trained for 20 epochs based on surrogate gradient. Returns the best-performing phenotype $Model$.}
    \STATE {Train $Model$ with surrogate gradients for 600 epochs.}
    \STATE {$Accuracy$ = Test($Model$)}
    \end{algorithmic}
  \end{algorithm}
\end{figure}

\subsubsection{Fitness Evaluation}
\begin{figure}[htp]
\centering
\includegraphics[width=8.5cm]{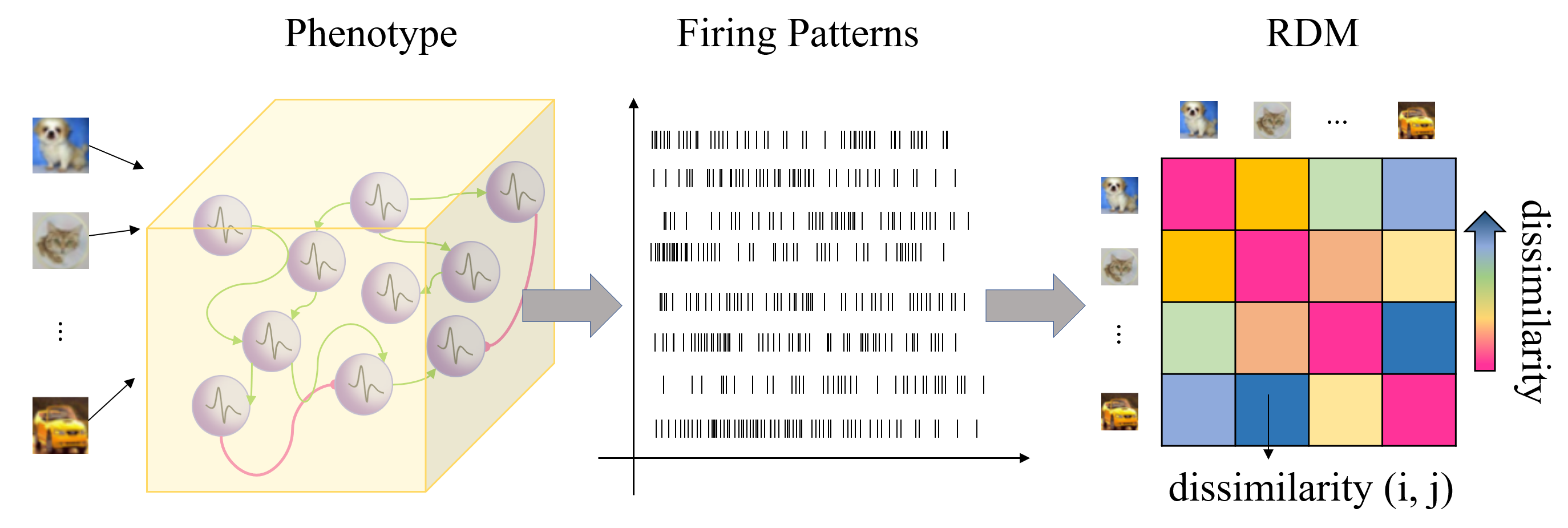}
\caption{BIE calculation process. Randomly select a batch of data from the dataset to input the decoded SNN model, and calculate BIE according to its firing patterns under different data.}
\label{select}
\end{figure}

Some neuroscience studies have found that brain function depends on stable activity and are less affected by different tasks~\cite{gratton2018functional,laumann2017stability}, whereas increased sensitivity to the environment typically results in reduced information processing capacity~\cite{sturman2011neurobiology,fuhrmann2015adolescence}.
To measure the efficiency of multi-motif collaborative SNN architecture, we design a brain-inspried indirect evaluation (BIE) to obtain more stable individuals by performing representation similarity analysis~\cite{kriegeskorte2008representational} on small batches of samples and exclude the negative effects of poor architecture on evolution.

The calculation process of the BIE is shown in Fig.~\ref{select}. Some samples are input into the decoded phenotype. Record the firing pattern $P_i$ of the genotype $Individual_i$ with $n_{neu}$ neurons. It is a $n_{neu}$ dimension vector, where each dimension represents the count of spikes of this neuron during the time period of $T$:
\begin{equation}
P_i=\left(\begin{array}{ccc}
\sum_{0}^T p_{1}^t & \cdots & \sum_{0}^T p_{n_{neu}}^t \\
\end{array}\right)\end{equation}

\begin{align}
\label{eq3}
p_z^t= \begin{cases}1 & \text { if the neuron $z$ fires at time t} \\ 0 & \text { if the neuron $z$ does not fire at time t}\end{cases}
\end{align}

The calculation method of BIE when inputting samples in batches is:
\begin{equation}
BIE_i=\left(\begin{array}{ccc}
d(P_i^1,P_i^1) & \cdots & d(P_i^1,P_i^j) \\
\vdots & \ddots & \vdots \\
d(P_i^j,P_i^1) & \cdots & d(P_i^j,P_i^j)
\end{array}\right)
\end{equation}
$P_i^j$ is the firing pattern of $Individual_i$ when inputting batches of samples (batch size is $j$). There are many ways to calculate the distance $d(u, v)$ (the dimension of $u$ and $v$ is $n$), after experimenting, we choose the Manhattan distance:
\begin{equation}
d(u, v)=\left|u_1-v_1\right|+\left|u_2-v_2\right|+...+\left|u_n-v_n\right|
\label{distance}
\end{equation}

\begin{table*}[htbp]
  \centering
  \caption{Comparison of classification performance on static datasets. The abbreviated training methods involved in the table: Spatio-Temporal Backpropagation (STBP)~\cite{wu2018spatio}; STBP NeuNorm~\cite{wu2019direct}; Temporal Spike Sequence Learning Backpropagation (TSSL-BP)~\cite{zhang2020temporal}; threshold-dependent Batch Normalization (tdBN)~\cite{zheng2021going}; Temporal Efficient Training (TET)~\cite{deng2022temporal}; Discrete Cosine Transform (DCT)~\cite{garg2021dct}; Time-To-First-Spike (TTFS)~\cite{park2020t2fsnn}.}
  \resizebox{6 in}{!}{
    \begin{tabular}{clcccc}
    \toprule
    \textbf{Dataset} & \textbf{Model} & \textbf{Training Methods} & \textbf{Architecture} & \textbf{Timesteps} & \textbf{Accuracy (\%)} \\
    \midrule
    \multirow{16}[4]{*}{CIFAR10} 
      & Wu et al.~\cite{wu2018spatio} & STBP & CIFARNet &12 & 89.83 \\
      & Wu et al.~\cite{wu2019direct} & STBP NeuNorm & CIFARNet &12 & 90.53 \\
      & Kundu et al.~\cite{kundu2021spike} &Hybrid &VGG16 &100 &91.29\\
      & Zhang \& Li ~\cite{zhang2020temporal} & TSSL-BP & CIFARNet &5 & 91.41 \\
      & Shen et aL ~\cite{shen2022backpropagation} & STBP & 7-layer-CNN &8 & 92.15 \\
      & Rathi et al.~\cite{rathi2020enabling} & Hybrid & ResNet-20 &250 & 92.22 \\
      & Rathi \& Roy ~\cite{rathi2020diet} & Diet-SNN & ResNet-20 &10 & 92.54 \\
      & Li et al.~\cite{li2021free} &ANN-SNN Conversion &VGG16 &32 &93.00\\
      & Zheng et al. ~\cite{zheng2021going} & STBP-tdBN & ResNet-19 &6 & 93.16 \\
      &Fang et al. ~\cite{fang2021incorporating}&STBP &6Conv, 2Linear &8 &93.50\\
      & Han et al.~\cite{han2020rmp}&ANN-SNN Conversion &VGG16 &2048 &93.63\\
      & Deng et al. ~\cite{deng2022temporal} & TET & ResNet-19 &4 & 94.44 \\
\cmidrule{2-6}      & Kim et al. ~\cite{kim2022neural} & STBP & NAS &5 & 92.73 \\
      & Na et al. ~\cite{na2022autosnn} & STBP & NAS &16 & 93.15 \\
    & \textbf{Our Method} & STBP NeuNorm & MSE-NAS &\textbf{4} & \textbf{96.19} \\

      & \textbf{Our Method} & STBP & MSE-NAS &\textbf{4} & \textbf{96.58} \\

    \midrule
    \multirow{14}[4]{*}{CIFAR100} 
    &Lu and Sengupta ~\cite{lu2020exploring}&ANN-SNN Conversion &VGG15 &62 &63.20\\
    & Rathi \& Roy ~\cite{rathi2020diet} & Diet-SNN & ResNet-20 &5 & 64.07 \\
    &Kundu et al.~\cite{kundu2021spike}&Hybrid &VGG11 &120 &64.98\\
    & Rathi et al. ~\cite{rathi2020enabling} & Hybrid & VGG-11 &125 & 67.87 \\
    &Garg et al.~\cite{garg2021dct} &DCT &VGG9 &48 &68.30\\
    &Deng et al. ~\cite{deng2021optimal}&ANN-SNN Conversion &ResNet-20  &32 &68.40\\
    & Park et al. ~\cite{park2020t2fsnn}& TTFS &VGG15 &680 &68.80\\
    & Shen et al. ~\cite{shen2022backpropagation} & STBP & ResNet34 &8 & 69.32 \\
    &Han et al. .~\cite{han2020rmp}&ANN-SNN Conversion &VGG16 &2048 &70.90\\
    &Li et al.~\cite{li2021free}  &ANN-SNN Conversion &ResNet-20  &16 &72.33\\
\cmidrule{2-6}   & Na et al. ~\cite{na2022autosnn} & STBP & NAS &16 & 69.16 \\
        & Kim et al.  ~\cite{kim2022neural} & STBP & NAS &5 & 73.04 \\
      & \textbf{Our Method} & STBP NeuNorm& MSE-NAS &\textbf{4} & \textbf{80.25} \\
    & \textbf{Our Method} & STBP  & MSE-NAS &\textbf{4} & \textbf{80.56} \\

    \bottomrule
    \end{tabular}}
  \label{stat}%
\end{table*}%

The evolution direction of genotype $i$ is set to minimize the difference in the response of the same individual to different samples. The objective problem of evolution can be expressed as follows:
\begin{equation}
\mathop{\arg\min}\limits_{[X_i,M_i,G_i]}\frac{1}{s}\sum_{j=1}^s BIE_i^j
\label{fit}
\end{equation}

 $s$ is the number of batches used for evaluation.

\begin{table*}[htbp]
  \centering
  \caption{Comparison of classification performance of MSE-NAS on neuromorphic datasets. The abbreviated training methods involved in the table: threshold-dependent Batch Normalization (tdBN)~\cite{zheng2021going}; Backpropagation Through
Time (BPTT)~\cite{werbos1988generalization}; Spatio-Temporal Backpropagation (STBP)~\cite{wu2018spatio}; Temporal Efficient Training (TET)~\cite{deng2022temporal};  Spike Layer Error Reassignment (SLAYER)~\cite{liu2001spike}.}
    \resizebox{6 in}{!}{\begin{tabular}{clcccc}
    \toprule
    \textbf{Dataset} & \textbf{Model} & \textbf{Training Methods} & \textbf{Architecture} & \textbf{Timesteps} & \textbf{Accuracy (\%)} \\
    \midrule
    \multirow{7}[4]{*}{CIFAR10-DVS}  & Kugele et al. ~\cite{kugele2020efficient}& Streaming Rollout & DenseNet &10 & 66.8 \\
    & Zheng et al. ~\cite{zheng2021going} & STBP-tdBN & ResNet-19 &10 & 67.8 \\
     & Wu et al. ~\cite{wu2021liaf} & BPTT & LIAF-Net &10 & 70.40 \\
      & Shen et al. ~\cite{shen2022backpropagation} & STBP & 5-layer-CNN &16 & 78.95 \\
      & Deng et al.  ~\cite{deng2022temporal} & TET & VGGSNN &10 & 83.17 \\
\cmidrule{2-6}      & Na et al. ~\cite{na2022autosnn} & STBP & NAS &16 & 72.50 \\
      & \textbf{Our Method} & STBP & MSE-NAS &\textbf{10} & \textbf{84.0} \\
    \midrule
    \multirow{5}[2]{*}{DVS128-Gesture} & Xing et al ~\cite{xing2020new} & SLAYER & 5-Layer-CNN &20 & 92.01 \\
      & Shrestha et al ~\cite{shrestha2018slayer} & SLAYER & 16-layer-CNN &300 & 93.64 \\
        & Fang et al.~\cite{fang2021incorporating} & STBP & 5Conv, 2Linear &20 & 97.57 \\
\cmidrule{2-6}      & Na et al. ~\cite{na2022autosnn} & STBP & NAS &16 & 96.53 \\

      & \textbf{Our Method} & STBP & MSE-NAS &\textbf{10} & \textbf{98.10} \\
    \bottomrule
    \end{tabular}}
  \label{neuro}%
\end{table*}%

\begin{table}[htp]

\begin{tabular}{llll}
    \toprule
 Dataset & Ablation Space & Description &Accuracy (\%) \\
    \midrule
 \multirow{8}{*}{CIFAR10} 
 
 &\multirow{6}{*}{Meso} 
 &MI& $92.63\pm 0.25$\\
&  &FE&$93.81\pm 0.64$\\
&  &FI&$94.46\pm 0.13$\\
&  & FbI&$95.29\pm 0.26$\\
&  &LI&$95.83\pm 0.19$\\
&  & \textbf{MSE-NAS}& $\bm{96.4\pm 0.12}$\\
\cmidrule{2-4}
 &\multirow{2}{*}{Macro} 
 &CL-0 & $93.55 \pm 0.3$\\
&  &\textbf{MSE-NAS}&$\bm{95.35 \pm 0.32}$\\

    \midrule
 \multirow{8}{*}{CIFAR100} 
 
 &\multirow{6}{*}{Meso} 
 &MI&$74.41\pm0.4$\\
&  &FI&$74.73\pm0.08$\\
&  &FE&$76.93\pm0.04$\\
&  & FbI&$76.96\pm0.04$\\
&  &LI&$77.56\pm0.22$\\
&  &\textbf{MSE-NAS}&$\bm{78.51\pm0.41}$\\
\cmidrule{2-4}
 &\multirow{2}{*}{Macro} 
 &CL-0 &$75.64\pm0.22$\\
&  &\textbf{MSE-NAS}&$\bm{76.5\pm0.12}$\\

    \midrule
 \multirow{8}{*}{CIFAR10-DVS} 
 
 &\multirow{6}{*}{Meso} 
 &MI&$79.89\pm0.1$\\
 &  & FbI&$80.0\pm0.8$\\
&  &FE&$80.3\pm0.3$\\
&  &FI&$81.1\pm0.3$\\
&  &LI&$81.35\pm0.6$\\
&  &\textbf{MSE-NAS}&$\bm{82.4\pm0.2}$\\
\cmidrule{2-4}
 &\multirow{2}{*}{Macro} 
 &CL-0 & $80.52\pm 0.52$\\
&  &\textbf{MSE-NAS}&$\bm{82.5\pm 0.1}$\\

 \midrule
  \multirow{8}{*}{DVS128-Gesture} 
 &\multirow{6}{*}{Meso} 
 &MI&$ 95.07\pm0.76$\\
&  &FE&$95.26\pm0.4$\\
&  &FI&$95.57\pm0.5$\\
&  & FbI&$95.58\pm0.63$\\
&  &LI&$95.95\pm0.5$\\
&  &\textbf{MSE-NAS}&$\bm{97.70\pm0.25}$\\
\cmidrule{2-4}
 &\multirow{2}{*}{Macro} 
 &CL-0 & $95.26\pm0.57$\\
&  &\textbf{MSE-NAS}& $\bm{97.47\pm0.31}$\\
\bottomrule

\end{tabular}
\caption{Ablation experiments in multi-scale search spaces.}\label{ablatab}\centering
\end{table}

\section{EXPERIMENTS AND ANALYSIS}\label{exp}
In order to verify the effectiveness of the proposed MSE-NAS algorithm, we conducted experiments on multiple static datasets CIFAR10~\cite{krizhevsky2009learning}, CIFAR100~\cite{xu2015empirical} and neuromorphic datasets DVS128-Gesture~\cite{amir2017low}, CIFAR10-DVS~\cite{li2017cifar10}. 
The neuromorphic dataset CIFAR10-DVS converts static images from CIFAR10 into event streams by capturing oriented gradients generated by images of repeated closed-loop smooth (RCLS) motion via Dynamic Vision Sensors (DVS).
DVS128-Gesture is also an event stream dataset captured by a DVS, containing 1342 instances of 11 hand and arm gestures, each lasting about 6 seconds.
Static datasets provide a snapshot of information at a given time, while neuromorphic datasets leverage event-stream-based data captured by Dynamic Vision Sensors (DVS), providing a dynamic view of the environment and capturing changes and events in real-time with high temporal resolution.
Table~\ref{stat} and Table~\ref{neuro} list the classification performance comparison between MSE-NAS and other methods on static and neuromorphic datasets respectively. 

At the time of initialization, each individual is set to have a maximum depth of 11, that is, a maximum of 11 motifs form a deep spiking neural network in the form of ordinary feedforward or cross-layer connections, and the last motif will pass through a global pooling layer. The feature maps are finally classified by a fully connected classifier.

\subsection{Results}
Table~\ref{stat} and~\ref{neuro} show the testing classification accuracy with the best individual (evolved by MSE-NAS for 80 generations) after training for 600 epochs. From Table~\ref{stat}, for the static dataset CIFAR10, MSE-NAS is significantly better than the artificially designed SNN architecture:  the classification accuracy of MSE-NAS is at least 2.14\% higher than CIFARNet, ResNet, VGG or other convolutional structures. Compared with the existing SNN-based NAS algorithms~\cite{kim2022neural,na2022autosnn}, under the same STBP training method, our method achieves a 3.85\% and 3.43\% performance improvement with fewer timesteps (only 4), respectively. Similar conclusions can be found on the static dataset CIFAR100. MSE-NAS outperforms the artificially designed SNN model by 8.23\% to 17.36\% and outperforms other NAS methods~\cite{kim2022neural,na2022autosnn} by 11.4\% and 7.52\%. While achieving the highest performance, the proposed model takes the least number of timesteps.

As shown in Table~\ref{neuro}, on the neuromorphic dataset CIFAR10-DVS, MSE-NAS achieves 0.83\% to 7.2\% accuracy improvement on the fixed architecture model. It uses only fewer timesteps to achieve higher performance compared to other SNN-based NAS (even achieves 11.5\%  performance improvement). On the DVS128-Gesture dataset, MSE-NAS is 0.53\% to 6.09\% higher than the fixed architecture model, and 1.57\% higher than other NAS for SNN methods. Our simulation time is shorter than all other methods.

Combining all methods listed above in Table~\ref{stat} and ~\ref{neuro}, we found that the accuracy of existing SNN-NAS models does not exceed the artificially designed architecture. Our proposed algorithm significantly improves both performance and energy consumption, demonstrating the superiority of MSE-NAS over existing SNN-NAS methods. 
To the best of our knowledge, MSE-NAS demonstrates state-of-the-art performance on the aforementioned four datasets. More importantly, it means that MSE-NAS surpasses the experience of human experts on SNN and expands the boundary of SNN architecture optimization.

\subsection{Ablation Study}
In order to prove that the evolution of different scales has an effect on improving the classification accuracy of SNN, we conducted ablation experiments on static datasets (CIFAR10, CIFAR100) and neuromorphic datasets (DVS128-Gesture, CIFAR10-DVS). The initial channels are set to 48, 48, 16, and 16. The maximum depth is limited to 12.



\subsubsection{Meso-Scale}

Inspired by the multi-scale topological space of the human brain, MSE-NAS evolves the combination of multiple neural microcircuits. In order to explore the effectiveness of this combination (i.e. the mesoscopic evolution space), we conducted comparative experiments on five single-motif architectures. We fix the type of microcircuits (FE, FI, FbI, MI, LI) at each layer throughout the model, and only evolve at macroscopic and microscopic levels. After multiple evolutions, Table~\ref{ablatab} illustrates the top-1 accuracy on four datasets. It can be seen that under this experimental setup, the evolution of multiple motifs (labelled MSE-NAS) improves classification accuracy by 0.57\% to 1.75\% compared to the best performance of the single-motif model on four datasets.

Across all datasets, MI exhibits the poorest performance. This could arise from the structure of MI, where inhibitory connections substantially outnumber excitatory ones, resulting in extremely unbalanced dynamics of the SNN network.
Some studies believe that the human brain neuron activity pattern shows a scale-invariant neuron avalanche phenomenon, that is, the excitation-inhibition (E/I) shows a balance, called self-organized criticality~\cite{plenz2021self,cramer2020control,beggs2003neuronal}.
In order to continue to explore the impact of E/I balance on MSE-NAS, we counted the relationship between the motif combinations of individuals and their performance on the DVS128-Gesture dataset.

\begin{figure}[htp]
\centering
\includegraphics[width=8cm]{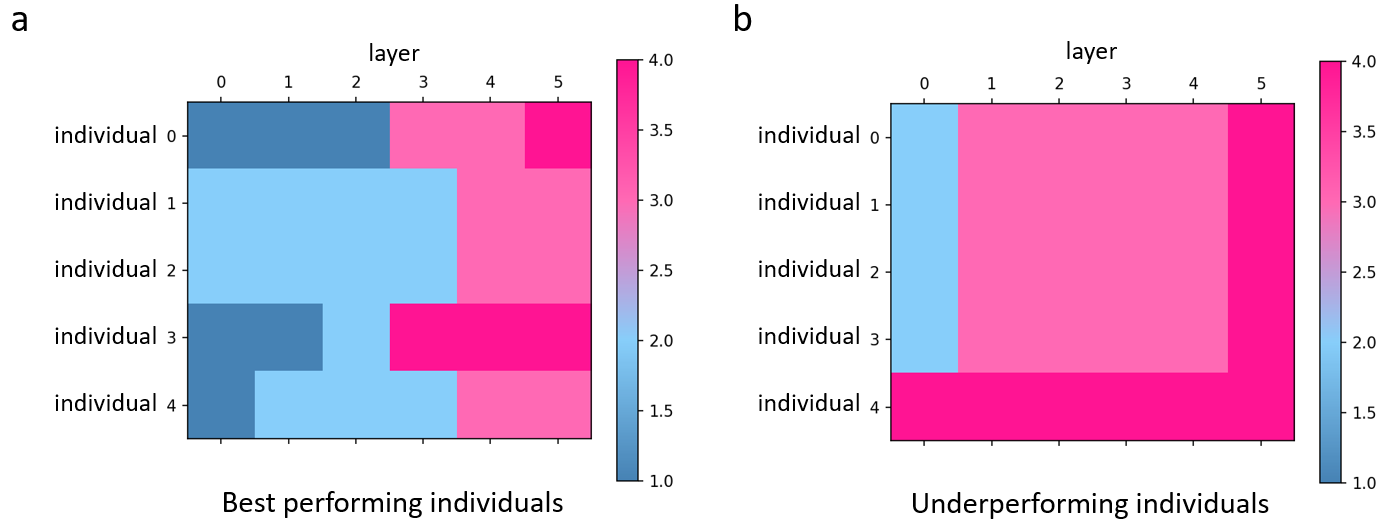}

\caption{Compare the effect of motif combination on individual classification accuracy. \textbf{a.} Individuals with high classification accuracy. \textbf{b.} Individuals with low classification accuracy.}
\label{top}
\end{figure}

Take the individuals with 6 layers as an example,
Fig.~\ref{top} shows two extreme cases: Fig.~\ref{top}a shows the motif combination of the five individuals with the highest classification accuracy; Fig.~\ref{top}b shows the motif combination of the five individuals with the lowest classification accuracy. We set four colours 1, 2, 3, 4, and the proportion of inhibitory neurons gradually increases from colour 1 to colour 4. The darker the blue, the more excitatory connections, the darker the red, the more inhibitory connections. 
FE, FI, FbI, LI, and MI are mapped to 1, 2, 2, 3, and 4, respectively, according to the ratio of excitatory-inhibitory connections among the five motifs. Since the proportion of excitatory-inhibitory neurons is the same in FbI and FI, we map them to the same colour 2. In Fig.~\ref{top}, the classification accuracy of individual 0 is the highest, and the classification accuracy of individual 5 is the lowest. Obviously, individuals with low performance contain more inhibitory modules, while individuals with high performance are a combination of multiple motifs. This fits with the equilibrium dynamics found in biological brains.



\subsubsection{Macro-Scale}

For the macro scale, the experiment compares whether the cross-motif connection is effective on multiple datasets. 
We fix the value of $g_1$ to 1 for MSE-NAS comparison. 
Table~\ref{ablatab} shows that on each dataset, the cross-layer connectivity (labelled MSE-NAS) is more efficient than the normal feed-forward structure (labelled CL-0), achieving 1.8\%, 0.9\%, 2.02\% and 2.21\% improvements on the four datasets, and the variance of classification accuracy is also smaller, as shown in Table~\ref{ablatab}. 
Cross-motif inputs provide motifs with richer feature information, indicating that the macro-scale evolution of SNNs inspired by long-term connections across brain regions in the nervous system is effective.

\subsection{Effects of Indirect Evaluation}
MSE-NAS employs a brain-inspired indirect method for evaluating individual fitness, thereby significantly reducing the time required for training individuals.
We test the time consumption on both static and neuromorphic datasets.
On CIFAR10, each individual undergoes training for 10 epochs by utilizing the direct evaluation method, which aids in stabilizing the training loss. Consequently, the evolution of 10 generations necessitates approximately 50 GPU hours. The indirect evaluation method without training takes only 0.83 GPU hours, making it approximately 60x faster. 
On DVS128-Gesture, it takes 14.4 GPU hours to evolve 10 generations using the direct evaluation method (training for 10 epochs to stabilize the loss), while the indirect evaluation method only needs 7.1 GPU hours, saving about half of the time.
Considering that CIFAR10 training takes longer time than DVS128-Gesture, an indirect evaluation approach would yield significant time-saving benefits. In essence, irrespective of the specific dataset utilized, indirect evaluation notably enhances the pace of the evolutionary process.

\begin{figure}[htp]
\centering
\includegraphics[width=7cm]{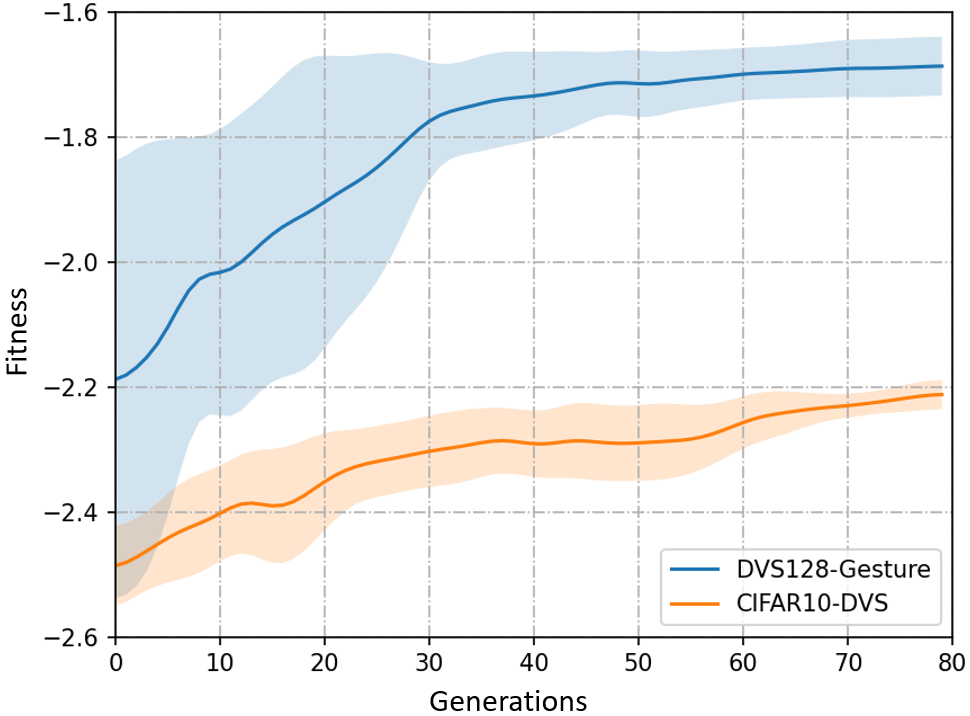}
\caption{The average fitness change of the population on DVS128-Gesture and CIFAR10-DVS.}
\label{accgens}
\end{figure}

Within 80 generations, the average fitness change of the population on DVS128-Gesture and CIFAR10-DVS are shown in Fig.~\ref{accgens}. Curves are the average fitness of the population per generation (multiple runs). As the evolution progresses, the average fitness of individuals gradually increases in both datasets.

\begin{figure*}[htp]
\centering
\includegraphics[width=17cm]{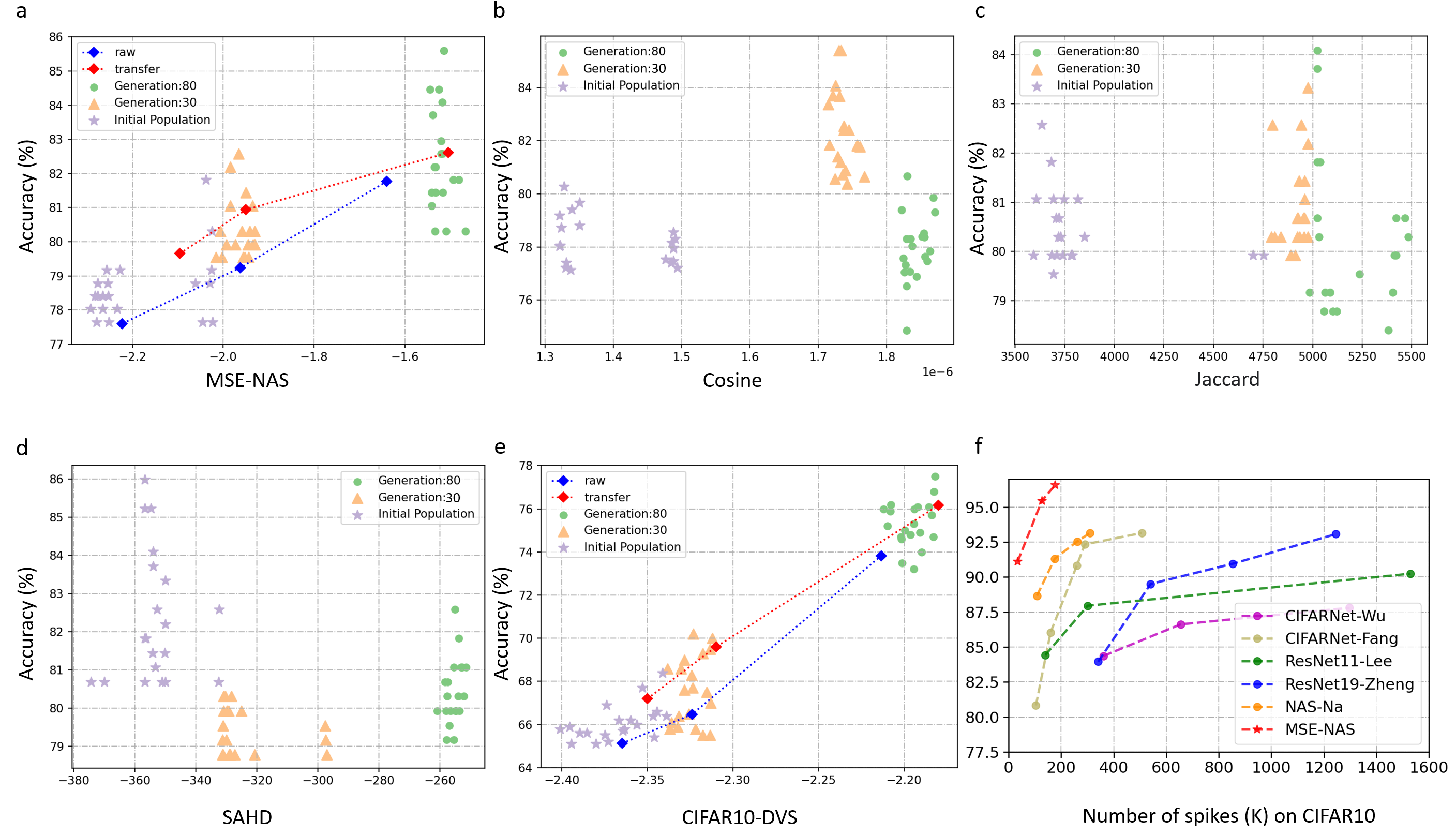}
\caption{Results of different evaluating methods and data transferring. \textbf{a.} Evolution and transferring results on DVS128-Gesture. \textbf{b.} Take cosine distance as the distance measurement. \textbf{c.} Take Jaccard distance as the distance measurement. \textbf{d.} Take SAHD as the fitness evaluation function. \textbf{e.}  Evolution and transferring results on CIFAR10-DVS. \textbf{f.} Comparison result of spike numbers of MSE-NAS with different precision on CIFAR10.}
\label{dis}
\end{figure*}
There are many distance metrics that can be used in \eqref{distance}, such as Jaccard distance~\eqref{jaccard} and cosine distance~\eqref{cosine} as the fitness function of evolution. In \eqref{jaccard}, $M_{00}$ represents the number of dimensions in which $x$ and $y$ are both 0, $M_{11}$ represents the number of dimensions in which $x$ and $y$ are both 1, $M_{10}$ represents the number of dimensions in which $x$ is 1 and $y$ is 0, $M_{01}$ represents the number of dimensions where the sample $x$ is 0 and $y$ is 1.

\begin{equation}
d_{Jac}(u, v)=\frac{M_{01}+M_{10}}{M_{01}+M_{10}+M_{11}}
\label{jaccard}
\end{equation}
\begin{equation}
d_{Cos}(u, v)=\frac{\mathrm{u}_1 \mathrm{v}_1+\mathrm{u}_2 \mathrm{v}_2+\cdots+\mathrm{u}_{n} \mathrm{v}_{n}}{\sqrt{\mathrm{u}_1^2+\cdots+\mathrm{u}_{n}^2} \cdot \sqrt{\mathrm{v}_1^2+\cdots+\mathrm{v}_{n}^2}}
\label{cosine}
\end{equation}

In addition, the SAHD indirect evaluation method proposed by ~\cite{kim2022neural} is also used to replace \eqref{fit} as a new fitness calculation method:
\begin{equation}
d_{S A H}\left(c_i^l, c_j^l\right)=\alpha d_H\left(c_i^l, c_j^l\right)
\end{equation}
where $\alpha$ is a desparse coefficient and $d_H$ is the Hamming distance metric. $c_i$ and $c_j$ are the activation patterns of different mini-batch samples $i$ and $j$.

We select the initial population, the 30th generation's population and the 80th generation's population as three time nodes, select the 20 individuals with the highest indirect fitness scores and train 20 epochs to obtain their classification accuracy. The scatter points indicate the classification accuracy and fitness of evolved individuals, and different colours represent different generations.
As the evolution proceeds, from the x-axis (fitness), in Fig.~\ref{dis}a-e, the value is constantly increasing from purple to orange to green, indicating that the fitness of the population is constantly improving.

The evolution of MSE-NAS on DVS128-Gesture and CIFAR10-DVS is shown in Fig.~\ref{dis}a and e respectively.
The blue lines represent the average performance of all individuals in each population.
From the initial generation, progressing through the 30th to the 80th generation, the increasing fitness has brought about a significant improvement in the top-1 accuracy, and such a positive correlation underscores the efficiency of the proposed evolutionary search algorithm.

As a comparison, models adopting the other three fitness functions fail to achieve improvement in classification performance, as shown in Fig.~\ref{dis}b (cosine), c (Jaccard) and d (SAHD).
We analyze that the proposed method (Fig.~\ref{dis}a and e) is effective because it directly pursues the stable-fired architecture: a good SNN architecture will not greatly increase or decrease the number of spikes due to the input samples. However, the response to different samples varies too much without training, which just shows that the architecture may have some preferences and is not robust enough.

\subsection{Transferability}
The human brain inherently possesses intricate cognitive functions and demonstrates remarkable generalization capabilities. 
Inspired by this, we hope that the evolved architectures have the potential of transferability in different tasks, negating the need to design dedicated for specific tasks. Thus we explore whether evolution facilitates architectural transfer between multiple datasets. Take the transfer on CIFAR10-DVS as an example. We first evolve the population on DVS128-Gesture to the last generation, and then take the top 20 individuals with the best performance as the initial population of CIFAR10-DVS (if the number of initial individuals is insufficient, copy to the same population size), and then evolve.

We conduct related experiments on DVS128-Gesture and CIFAR10-DVS, as shown in the lines of Fig.~\ref{dis}a and Fig.~\ref{dis}e respectively. The red line in Fig.~\ref{dis}a is the individuals' performance transferred from CIFAR10-DVS, while the red line in Fig.~\ref{dis}e is the individuals' performance transferred from DVS128-Gesture. The blue lines are the performances of evolved individuals on DVS128-Gesture and CIFAR10-DVS, respectively. Observations show that when individuals who evolved on one dataset are transferred to a new one, they consistently exhibit high fitness and classification accuracy from the outset, maintaining their superiority throughout the evolutionary process. This underscores that the transferred individuals can swiftly adapt to new tasks and evolve superior architectures.

Since the indirect evaluation method is data-independent, it only focuses on the model's response to different inputs. Top-performing individuals evolved by MSE-NAS on one dataset enhance the evolution efficiency on another dataset, proving that our proposed method exhibits strong transferability between datasets.

\subsection{Robustness Analysis}
\subsubsection {Training Method}
To investigate the influence of the training method on the performance of the model,  we train the model evolved by MSE-NAS using NeuNorm, as proposed by~\cite{wu2019direct}, on CIFAR10. 
After training for 600 epochs,  the model achieves a top-1 accuracy of 96.19\%, marking an increase of 5.66\% from the 90.53\% reported in~\cite{wu2019direct}, and is just 0.39\% lower than the best-reported 96.58\% of MSE-NAS trained by STBP.
We do the same experiment on CIFAR100 and achieve the result of 80.25\%, which is 0.31\% lower than our reported best result as shown in Table~\ref{stat}. 
No matter which training method is adopted, MSE-NAS has achieved the effect of SOTA compared with other fixed or NAS-based SNN architectures.
This underscores the assertion that the efficacy of models evolved by MSE-NAS transcends specific training methods, showcasing notable generalization capabilities.

\subsubsection {Resilience To Noise}
We assess the robustness of MSE-NAS by introducing Gaussian noise to the model's input and comparing its performance against other techniques.
In line with the approach taken by~\cite{kim2023exploring} (denoted as TIC), we introduce a Gaussian noise perturbation with a norm equal to 50\% of the input's norm for each image, ensuring the noise remains independent across images.
The comparison results of the noise experiment are shown in Table~\ref{noise}. In the absence of noise interference, the top-1 accuracy of MSE-NAS is 4.54\% and 12.09\% higher than TIC on CIFAR10 and CIFAR100 respectively.
Noise leads to a reduction in the top-1 accuracy of MSE-NAS by 7.94\% on CIFAR10 and 17.23\% on CIFAR100. However, this decline is still less significant than the best results observed on TIC, which are 23.03\% and 30.57\% respectively.
The above results highlight the remarkable robustness of MSE-NAS by showcasing its resilience against random noise disturbances.

\begin{table}[htp]
\begin{tabular}{llll}
    \toprule
Dataset &Method& Clean Acc (\%) & Gaussian Noise (\%) \\
    \midrule

     \multirow{2}{*}{CIFAR10} 

& TIC  & 92.04 & $69.01(-23.03)$ \\
&MSE-NAS & $\mathbf{9 6 . 5 8}$ & $\mathbf{8 8 . 6 4}(\mathbf{- 7 . 9 4})$ \\
    \midrule

     \multirow{3}{*}{CIFAR100} 

& TIC & 68.17 & $37.60(-30.57)$ \\
&TIC  & 68.47 & $36.98(-31.49)$ \\
& MSE-NAS & $\mathbf{8 0 . 5 6}$ & $\mathbf{63.33}(\mathbf{- 17.23})$ \\
\bottomrule
\end{tabular}
\caption{Effect of Noise on Classification Performance. The top-1 accuracy of the model without noise is denoted as Clean Acc, while the result with noise is denoted as Gaussian Noise. }
\label{noise}
\end{table}

\subsection{Energy Consumption Analysis}

In SNN, signals propagate in the form of spikes, which exhibits event-driven characteristics closer to biological neurons and is more energy-efficient than traditional ANNs. We obtained MSE-NAS with different accuracy by changing the channels and network depth. Fig.~\ref{distance} depicts the accuracy versus the total number of spikes of different SNN models on CIFAR10. The comparison results show that the model evolved by NAS \cite{na2022autosnn} has fewer spikes and higher accuracy than SNNs with fixed architectures, and MSE-NAS is significantly better than fixed SNN architecture or NAS-based SNN architectures, achieving 96.58\% classification accuracy with the smallest number of spikes $177.6K$.

Compared with \cite{na2022autosnn} which also uses NAS to evolve the network structure (310k), MSE-NAS only needs about half of the spike consumption while improving the performance by 3.43\%. 
Under the same energy consumption, the performance of \cite{na2022autosnn} is 5.26\% lower than that of MSE-NAS. 
Compared with the SNN structure with fixed structure, such as \cite{fang2021incorporating}, the energy consumption of MSE-NAS is about 1/3 of it (507K), and the performance is improved by 3.43\%.

Overall, our model significantly reduces the cost of spikes on top of performance improvements, providing new perspectives for more energy-efficient spiking neural network architecture design.

\section{Conclusion and future work}\label{con}
Almost all previous studies of SNNs are based on the well-established and effective architecture of DNNs. Benefiting from the emergence of NAS,  a small number of works began to discuss the architecture optimization of SNN, but there is still a lot of room for improvement in performance and brain-inspired mechanisms.
Inspired by the architecture evolution mechanism in the biological nervous system, this work constructs a multi-scale evolutionary spiking neural network search algorithm (MSE-NAS), which takes into account individual neurons, motifs topology, as well as long-term connectivity evolution. We also propose an indirect evaluation method for individuals to improve the efficiency of evolution without training.

Experiments confirm that the proposed MSE-NAS achieves superior performance on multiple datasets compared with the fixed SNN architecture and other existing SNN-based NAS models. We discover the excitatory-inhibitory balance phenomenon of the brain in the evolutionary product and verified the effectiveness of the brain-inspired multi-motif fusion SNN architecture on classification tasks.
Whether it is performance or energy consumption, MSE-NAS surpasses other fixed architectures or evolved SNN models. MSE-NAS can transfer on different datasets, which can help improve evolution efficiency and possess generalization. Moreover, the performance of MSE-NAS is not affected by the training method and noise, showing strong robustness.

The connections in the brain are intricate, and with the advancement of neuroscience, more and more neurocognitive processing mechanisms can be used to optimize the architectural search space of SNNs. We look forward to drawing more inspiration from biological neural data for the research of ENAS in the future. Due to its data-independent nature, the indirect evaluation method used by MSE-NAS can be applied to the exploration of transfer learning, continual learning, and multi-objective optimization problems in the future, providing a new perspective and helping for the evolution of more adaptive spiking neural network architecture and more general artificial intelligence.

\section{Acknowledgement}
This work is supported by the National Key Research and Development Program (Grant No. 2020AAA0107800),  the National Natural Science Foundation of China (Grant No. 62106261).

\bibliography{IEEEabrv,bibfile}

\begin{thebibliography}{100}
\providecommand{\url}[1]{#1}
\csname url@samestyle\endcsname
\providecommand{\newblock}{\relax}
\providecommand{\bibinfo}[2]{#2}
\providecommand{\BIBentrySTDinterwordspacing}{\spaceskip=0pt\relax}
\providecommand{\BIBentryALTinterwordstretchfactor}{4}
\providecommand{\BIBentryALTinterwordspacing}{\spaceskip=\fontdimen2\font plus
\BIBentryALTinterwordstretchfactor\fontdimen3\font minus
  \fontdimen4\font\relax}
\providecommand{\BIBforeignlanguage}[2]{{%
\expandafter\ifx\csname l@#1\endcsname\relax
\typeout{** WARNING: IEEEtran.bst: No hyphenation pattern has been}%
\typeout{** loaded for the language `#1'. Using the pattern for}%
\typeout{** the default language instead.}%
\else
\language=\csname l@#1\endcsname
\fi
#2}}
\providecommand{\BIBdecl}{\relax}
\BIBdecl

\bibitem{luo2021architectures}
L.~Luo, ``Architectures of neuronal circuits,'' \emph{Science}, vol. 373, no.
  6559, p. eabg7285, 2021.

\bibitem{pan2020activity}
Y.~Pan and M.~Monje, ``Activity shapes neural circuit form and function: a
  historical perspective,'' \emph{Journal of Neuroscience}, vol.~40, no.~5, pp.
  944--954, 2020.

\bibitem{isaacson2011inhibition}
J.~S. Isaacson and M.~Scanziani, ``How inhibition shapes cortical activity,''
  \emph{Neuron}, vol.~72, no.~2, pp. 231--243, 2011.

\bibitem{maass1997networks}
W.~Maass, ``Networks of spiking neurons: the third generation of neural network
  models,'' \emph{Neural networks}, vol.~10, no.~9, pp. 1659--1671, 1997.

\bibitem{diehl2015fast}
P.~U. Diehl, D.~Neil, J.~Binas, M.~Cook, S.-C. Liu, and M.~Pfeiffer,
  ``Fast-classifying, high-accuracy spiking deep networks through weight and
  threshold balancing,'' in \emph{2015 International joint conference on neural
  networks (IJCNN)}.\hskip 1em plus 0.5em minus 0.4em\relax ieee, 2015, pp.
  1--8.

\bibitem{lee2016training}
J.~H. Lee, T.~Delbruck, and M.~Pfeiffer, ``Training deep spiking neural
  networks using backpropagation,'' \emph{Frontiers in neuroscience}, vol.~10,
  p. 508, 2016.

\bibitem{lee2020enabling}
C.~Lee, S.~S. Sarwar, P.~Panda, G.~Srinivasan, and K.~Roy, ``Enabling
  spike-based backpropagation for training deep neural network architectures,''
  \emph{Frontiers in neuroscience}, p. 119, 2020.

\bibitem{gu2019stca}
P.~Gu, R.~Xiao, G.~Pan, and H.~Tang, ``Stca: Spatio-temporal credit assignment
  with delayed feedback in deep spiking neural networks.'' in \emph{IJCAI},
  2019, pp. 1366--1372.

\bibitem{wu2021training}
H.~Wu, Y.~Zhang, W.~Weng, Y.~Zhang, Z.~Xiong, Z.-J. Zha, X.~Sun, and F.~Wu,
  ``Training spiking neural networks with accumulated spiking flow,'' in
  \emph{Proceedings of the AAAI conference on artificial intelligence},
  vol.~35, no.~12, 2021, pp. 10\,320--10\,328.

\bibitem{wu2018spatio}
Y.~Wu, L.~Deng, G.~Li, J.~Zhu, and L.~Shi, ``Spatio-temporal backpropagation
  for training high-performance spiking neural networks,'' \emph{Frontiers in
  neuroscience}, vol.~12, p. 331, 2018.

\bibitem{he2016deep}
K.~He, X.~Zhang, S.~Ren, and J.~Sun, ``Deep residual learning for image
  recognition,'' in \emph{Proceedings of the IEEE conference on computer vision
  and pattern recognition}, 2016, pp. 770--778.

\bibitem{simonyan2014very}
K.~Simonyan and A.~Zisserman, ``Very deep convolutional networks for
  large-scale image recognition,'' \emph{arXiv preprint arXiv:1409.1556}, 2014.

\bibitem{vaswani2017attention}
A.~Vaswani, N.~Shazeer, N.~Parmar, J.~Uszkoreit, L.~Jones, A.~N. Gomez,
  {\L}.~Kaiser, and I.~Polosukhin, ``Attention is all you need,''
  \emph{Advances in neural information processing systems}, vol.~30, 2017.

\bibitem{sengupta2019going}
A.~Sengupta, Y.~Ye, R.~Wang, C.~Liu, and K.~Roy, ``Going deeper in spiking
  neural networks: Vgg and residual architectures,'' \emph{Frontiers in
  neuroscience}, vol.~13, p.~95, 2019.

\bibitem{szegedy2015going}
C.~Szegedy, W.~Liu, Y.~Jia, P.~Sermanet, S.~Reed, D.~Anguelov, D.~Erhan,
  V.~Vanhoucke, and A.~Rabinovich, ``Going deeper with convolutions,'' in
  \emph{Proceedings of the IEEE conference on computer vision and pattern
  recognition}, 2015, pp. 1--9.

\bibitem{maass2002real}
W.~Maass, T.~Natschl{\"a}ger, and H.~Markram, ``Real-time computing without
  stable states: A new framework for neural computation based on
  perturbations,'' \emph{Neural computation}, vol.~14, no.~11, pp. 2531--2560,
  2002.

\bibitem{suarez2021learning}
L.~E. Su{\'a}rez, B.~A. Richards, G.~Lajoie, and B.~Misic, ``Learning function
  from structure in neuromorphic networks,'' \emph{Nature Machine
  Intelligence}, vol.~3, no.~9, pp. 771--786, 2021.

\bibitem{pham2018efficient}
H.~Pham, M.~Guan, B.~Zoph, Q.~Le, and J.~Dean, ``Efficient neural architecture
  search via parameters sharing,'' in \emph{International conference on machine
  learning}.\hskip 1em plus 0.5em minus 0.4em\relax PMLR, 2018, pp. 4095--4104.

\bibitem{liu2018darts}
H.~Liu, K.~Simonyan, and Y.~Yang, ``Darts: Differentiable architecture
  search,'' \emph{arXiv preprint arXiv:1806.09055}, 2018.

\bibitem{zela2019understanding}
A.~Zela, T.~Elsken, T.~Saikia, Y.~Marrakchi, T.~Brox, and F.~Hutter,
  ``Understanding and robustifying differentiable architecture search,''
  \emph{arXiv preprint arXiv:1909.09656}, 2019.

\bibitem{zoph2016neural}
B.~Zoph and Q.~V. Le, ``Neural architecture search with reinforcement
  learning,'' \emph{arXiv preprint arXiv:1611.01578}, 2016.

\bibitem{zoph2018learning}
B.~Zoph, V.~Vasudevan, J.~Shlens, and Q.~V. Le, ``Learning transferable
  architectures for scalable image recognition,'' in \emph{Proceedings of the
  IEEE conference on computer vision and pattern recognition}, 2018, pp.
  8697--8710.

\bibitem{baker2016designing}
B.~Baker, O.~Gupta, N.~Naik, and R.~Raskar, ``Designing neural network
  architectures using reinforcement learning,'' \emph{arXiv preprint
  arXiv:1611.02167}, 2016.

\bibitem{real2017large}
E.~Real, S.~Moore, A.~Selle, S.~Saxena, Y.~L. Suematsu, J.~Tan, Q.~V. Le, and
  A.~Kurakin, ``Large-scale evolution of image classifiers,'' in
  \emph{International Conference on Machine Learning}.\hskip 1em plus 0.5em
  minus 0.4em\relax PMLR, 2017, pp. 2902--2911.

\bibitem{xie2017genetic}
L.~Xie and A.~Yuille, ``Genetic cnn,'' in \emph{Proceedings of the IEEE
  international conference on computer vision}, 2017, pp. 1379--1388.

\bibitem{wen2021two}
Y.-W. Wen, S.-H. Peng, and C.-K. Ting, ``Two-stage evolutionary neural
  architecture search for transfer learning,'' \emph{IEEE Transactions on
  Evolutionary Computation}, vol.~25, no.~5, pp. 928--940, 2021.

\bibitem{floreano2008neuroevolution}
D.~Floreano, P.~D{\"u}rr, and C.~Mattiussi, ``Neuroevolution: from
  architectures to learning,'' \emph{Evolutionary intelligence}, vol.~1, pp.
  47--62, 2008.

\bibitem{liu2021survey}
Y.~Liu, Y.~Sun, B.~Xue, M.~Zhang, G.~G. Yen, and K.~C. Tan, ``A survey on
  evolutionary neural architecture search,'' \emph{IEEE transactions on neural
  networks and learning systems}, 2021.

\bibitem{zhu2021real}
H.~Zhu and Y.~Jin, ``Real-time federated evolutionary neural architecture
  search,'' \emph{IEEE Transactions on Evolutionary Computation}, vol.~26,
  no.~2, pp. 364--378, 2021.

\bibitem{sun2019evolving}
Y.~Sun, B.~Xue, M.~Zhang, and G.~G. Yen, \emph{IEEE Transactions on
  Evolutionary Computation}, vol.~24, no.~2, pp. 394--407, 2019.

\bibitem{zhang2022evolutionary}
H.~Zhang, Y.~Jin, and K.~Hao, ``Evolutionary search for complete neural network
  architectures with partial weight sharing,'' \emph{IEEE Transactions on
  Evolutionary Computation}, vol.~26, no.~5, pp. 1072--1086, 2022.

\bibitem{xie2022benchenas}
X.~Xie, Y.~Liu, Y.~Sun, G.~G. Yen, B.~Xue, and M.~Zhang, ``Benchenas: A
  benchmarking platform for evolutionary neural architecture search,''
  \emph{IEEE Transactions on Evolutionary Computation}, vol.~26, no.~6, pp.
  1473--1485, 2022.

\bibitem{stanley2019designing}
K.~O. Stanley, J.~Clune, J.~Lehman, and R.~Miikkulainen, ``Designing neural
  networks through neuroevolution,'' \emph{Nature Machine Intelligence},
  vol.~1, no.~1, pp. 24--35, 2019.

\bibitem{stockl2021structure}
C.~St{\"o}ckl, D.~Lang, and W.~Maass, ``Structure induces computational
  function in networks with diverse types of spiking neurons,'' \emph{bioRxiv},
  pp. 2021--05, 2021.

\bibitem{na2022autosnn}
B.~Na, J.~Mok, S.~Park, D.~Lee, H.~Choe, and S.~Yoon, ``Autosnn: towards
  energy-efficient spiking neural networks,'' in \emph{International Conference
  on Machine Learning}.\hskip 1em plus 0.5em minus 0.4em\relax PMLR, 2022, pp.
  16\,253--16\,269.

\bibitem{kim2022neural}
Y.~Kim, Y.~Li, H.~Park, Y.~Venkatesha, and P.~Panda, ``Neural architecture
  search for spiking neural networks,'' in \emph{Computer Vision--ECCV 2022:
  17th European Conference, Tel Aviv, Israel, October 23--27, 2022,
  Proceedings, Part XXIV}.\hskip 1em plus 0.5em minus 0.4em\relax Springer,
  2022, pp. 36--56.

\bibitem{kriegeskorte2008representational}
N.~Kriegeskorte, M.~Mur, and P.~A. Bandettini, ``Representational similarity
  analysis-connecting the branches of systems neuroscience,'' \emph{Frontiers
  in systems neuroscience}, p.~4, 2008.

\bibitem{maass2004methods}
W.~Maass, R.~Legenstein, and N.~Bertschinger, ``Methods for estimating the
  computational power and generalization capability of neural microcircuits,''
  \emph{Advances in neural information processing systems}, vol.~17, 2004.

\bibitem{liu2019auto}
C.~Liu, L.-C. Chen, F.~Schroff, H.~Adam, W.~Hua, A.~L. Yuille, and L.~Fei-Fei,
  ``Auto-deeplab: Hierarchical neural architecture search for semantic image
  segmentation,'' in \emph{Proceedings of the IEEE/CVF conference on computer
  vision and pattern recognition}, 2019, pp. 82--92.

\bibitem{zhang2019customizable}
Y.~Zhang, Z.~Qiu, J.~Liu, T.~Yao, D.~Liu, and T.~Mei, ``Customizable
  architecture search for semantic segmentation,'' in \emph{Proceedings of the
  IEEE/CVF Conference on Computer Vision and Pattern Recognition}, 2019, pp.
  11\,641--11\,650.

\bibitem{ghiasi2019fpn}
G.~Ghiasi, T.-Y. Lin, and Q.~V. Le, ``Nas-fpn: Learning scalable feature
  pyramid architecture for object detection,'' in \emph{Proceedings of the
  IEEE/CVF conference on computer vision and pattern recognition}, 2019, pp.
  7036--7045.

\bibitem{li2019partial}
X.~Li, Y.~Zhou, Z.~Pan, and J.~Feng, ``Partial order pruning: for best
  speed/accuracy trade-off in neural architecture search,'' in
  \emph{Proceedings of the IEEE/CVF Conference on Computer Vision and Pattern
  Recognition}, 2019, pp. 9145--9153.

\bibitem{fedorov2019sparse}
I.~Fedorov, R.~P. Adams, M.~Mattina, and P.~Whatmough, ``Sparse: Sparse
  architecture search for cnns on resource-constrained microcontrollers,''
  \emph{Advances in Neural Information Processing Systems}, vol.~32, 2019.

\bibitem{wistuba2019deep}
M.~Wistuba, ``Deep learning architecture search by neuro-cell-based evolution
  with function-preserving mutations,'' in \emph{Machine Learning and Knowledge
  Discovery in Databases: European Conference, ECML PKDD 2018, Dublin, Ireland,
  September 10--14, 2018, Proceedings, Part II 18}.\hskip 1em plus 0.5em minus
  0.4em\relax Springer, 2019, pp. 243--258.

\bibitem{elsken2017simple}
T.~Elsken, J.-H. Metzen, and F.~Hutter, ``Simple and efficient architecture
  search for convolutional neural networks,'' \emph{arXiv preprint
  arXiv:1711.04528}.

\bibitem{chen2019progressive}
X.~Chen, L.~Xie, J.~Wu, and Q.~Tian, ``Progressive differentiable architecture
  search: Bridging the depth gap between search and evaluation,'' in
  \emph{Proceedings of the IEEE/CVF international conference on computer
  vision}, 2019, pp. 1294--1303.

\bibitem{zhong2018practical}
Z.~Zhong, J.~Yan, W.~Wu, J.~Shao, and C.-L. Liu, ``Practical block-wise neural
  network architecture generation,'' in \emph{Proceedings of the IEEE
  conference on computer vision and pattern recognition}, 2018, pp. 2423--2432.

\bibitem{liu2017hierarchical}
H.~Liu, K.~Simonyan, O.~Vinyals, C.~Fernando, and K.~Kavukcuoglu,
  ``Hierarchical representations for efficient architecture search,''
  \emph{arXiv preprint arXiv:1711.00436}, 2017.

\bibitem{dong2018dpp}
J.-D. Dong, A.-C. Cheng, D.-C. Juan, W.~Wei, and M.~Sun, ``Dpp-net:
  Device-aware progressive search for pareto-optimal neural architectures,'' in
  \emph{Proceedings of the European Conference on Computer Vision (ECCV)},
  2018, pp. 517--531.

\bibitem{cui2019fast}
J.~Cui, P.~Chen, R.~Li, S.~Liu, X.~Shen, and J.~Jia, ``Fast and practical
  neural architecture search,'' in \emph{Proceedings of the IEEE/CVF
  International Conference on Computer Vision}, 2019, pp. 6509--6518.

\bibitem{wu2019fbnet}
B.~Wu, X.~Dai, P.~Zhang, Y.~Wang, F.~Sun, Y.~Wu, Y.~Tian, P.~Vajda, Y.~Jia, and
  K.~Keutzer, ``Fbnet: Hardware-aware efficient convnet design via
  differentiable neural architecture search,'' in \emph{Proceedings of the
  IEEE/CVF Conference on Computer Vision and Pattern Recognition}, 2019, pp.
  10\,734--10\,742.

\bibitem{huang2017densely}
G.~Huang, Z.~Liu, L.~Van Der~Maaten, and K.~Q. Weinberger, ``Densely connected
  convolutional networks,'' in \emph{Proceedings of the IEEE conference on
  computer vision and pattern recognition}, 2017, pp. 4700--4708.

\bibitem{howard2019searching}
A.~Howard, M.~Sandler, G.~Chu, L.-C. Chen, B.~Chen, M.~Tan, W.~Wang, Y.~Zhu,
  R.~Pang, V.~Vasudevan \emph{et~al.}, ``Searching for mobilenetv3,'' in
  \emph{Proceedings of the IEEE/CVF international conference on computer
  vision}, 2019, pp. 1314--1324.

\bibitem{chen2019auto}
Z.~Chen, Y.~Zhou, and Z.~Huang, ``Auto-creation of effective neural network
  architecture by evolutionary algorithm and resnet for image classification,''
  in \emph{2019 IEEE international conference on systems, man and cybernetics
  (SMC)}.\hskip 1em plus 0.5em minus 0.4em\relax IEEE, 2019, pp. 3895--3900.

\bibitem{song2020efficient}
D.~Song, C.~Xu, X.~Jia, Y.~Chen, C.~Xu, and Y.~Wang, ``Efficient residual dense
  block search for image super-resolution,'' in \emph{Proceedings of the AAAI
  Conference on Artificial Intelligence}, vol.~34, no.~07, 2020, pp.
  12\,007--12\,014.

\bibitem{sun2019surrogate}
Y.~Sun, H.~Wang, B.~Xue, Y.~Jin, G.~G. Yen, and M.~Zhang, ``Surrogate-assisted
  evolutionary deep learning using an end-to-end random forest-based
  performance predictor,'' \emph{IEEE Transactions on Evolutionary
  Computation}, vol.~24, no.~2, pp. 350--364, 2019.

\bibitem{abdelfattah2021zero}
M.~S. Abdelfattah, A.~Mehrotra, {\L}.~Dudziak, and N.~D. Lane, ``Zero-cost
  proxies for lightweight nas,'' \emph{arXiv preprint arXiv:2101.08134}, 2021.

\bibitem{lapicque1907recherches}
L.~Lapicque, ``Recherches quantitatives sur l'excitation electrique des nerfs
  traitee comme une polarization,'' \emph{Journal de physiologie et de
  pathologie g{\'e}n{\'e}rale}, vol.~9, pp. 620--635, 1907.

\bibitem{zeng2022braincog}
Y.~Zeng, D.~Zhao, F.~Zhao, G.~Shen, Y.~Dong, E.~Lu, Q.~Zhang, Y.~Sun, Q.~Liang,
  Y.~Zhao \emph{et~al.}, ``Braincog: A spiking neural network based
  brain-inspired cognitive intelligence engine for brain-inspired ai and brain
  simulation,'' \emph{arXiv preprint arXiv:2207.08533}, 2022.

\bibitem{meredith2002neuronal}
M.~A. Meredith, ``On the neuronal basis for multisensory convergence: a brief
  overview,'' \emph{Cognitive brain research}, vol.~14, no.~1, pp. 31--40,
  2002.

\bibitem{sherrington1925remarks}
C.~S. Sherrington, ``Remarks on some aspects of reflex inhibition,''
  \emph{Proceedings of the Royal Society of London. Series B, Containing Papers
  of a Biological Character}, vol.~97, no. 686, pp. 519--545, 1925.

\bibitem{coombs1955inhibitory}
J.~Coombs, J.~Eccles, and P.~Fatt, ``The inhibitory suppression of reflex
  discharges from motoneurones,'' \emph{The Journal of physiology}, vol. 130,
  no.~2, p. 396, 1955.

\bibitem{matlin1992sensation}
M.~W. Matlin and H.~J. Foley, \emph{Sensation and perception}.\hskip 1em plus
  0.5em minus 0.4em\relax Allyn \& Bacon, 1992.

\bibitem{orchard1991neural}
G.~A. Orchard and W.~A. Phillips, \emph{Neural computation: A beginner's
  guide}.\hskip 1em plus 0.5em minus 0.4em\relax Taylor \& Francis, 1991.

\bibitem{laurent2001odor}
G.~Laurent, M.~Stopfer, R.~W. Friedrich, M.~I. Rabinovich, A.~Volkovskii, and
  H.~D. Abarbanel, ``Odor encoding as an active, dynamical process:
  experiments, computation, and theory,'' \emph{Annual review of neuroscience},
  vol.~24, no.~1, pp. 263--297, 2001.

\bibitem{mountcastle1957modality}
V.~B. Mountcastle, ``Modality and topographic properties of single neurons of
  cat's somatic sensory cortex,'' \emph{Journal of neurophysiology}, vol.~20,
  no.~4, pp. 408--434, 1957.

\bibitem{grillner2006biological}
S.~Grillner, ``Biological pattern generation: the cellular and computational
  logic of networks in motion,'' \emph{Neuron}, vol.~52, no.~5, pp. 751--766,
  2006.

\bibitem{saper2010sleep}
C.~B. Saper, P.~M. Fuller, N.~P. Pedersen, J.~Lu, and T.~E. Scammell, ``Sleep
  state switching,'' \emph{Neuron}, vol.~68, no.~6, pp. 1023--1042, 2010.

\bibitem{gratton2018functional}
C.~Gratton, T.~O. Laumann, A.~N. Nielsen, D.~J. Greene, E.~M. Gordon, A.~W.
  Gilmore, S.~M. Nelson, R.~S. Coalson, A.~Z. Snyder, B.~L. Schlaggar
  \emph{et~al.}, ``Functional brain networks are dominated by stable group and
  individual factors, not cognitive or daily variation,'' \emph{Neuron},
  vol.~98, no.~2, pp. 439--452, 2018.

\bibitem{laumann2017stability}
T.~O. Laumann, A.~Z. Snyder, A.~Mitra, E.~M. Gordon, C.~Gratton, B.~Adeyemo,
  A.~W. Gilmore, S.~M. Nelson, J.~J. Berg, D.~J. Greene \emph{et~al.}, ``On the
  stability of bold fmri correlations,'' \emph{Cerebral cortex}, vol.~27,
  no.~10, pp. 4719--4732, 2017.

\bibitem{sturman2011neurobiology}
D.~A. Sturman and B.~Moghaddam, ``The neurobiology of adolescence: changes in
  brain architecture, functional dynamics, and behavioral tendencies,''
  \emph{Neuroscience \& Biobehavioral Reviews}, vol.~35, no.~8, pp. 1704--1712,
  2011.

\bibitem{fuhrmann2015adolescence}
D.~Fuhrmann, L.~J. Knoll, and S.-J. Blakemore, ``Adolescence as a sensitive
  period of brain development,'' \emph{Trends in cognitive sciences}, vol.~19,
  no.~10, pp. 558--566, 2015.

\bibitem{wu2019direct}
Y.~Wu, L.~Deng, G.~Li, J.~Zhu, Y.~Xie, and L.~Shi, ``Direct training for
  spiking neural networks: Faster, larger, better,'' in \emph{Proceedings of
  the AAAI conference on artificial intelligence}, vol.~33, no.~01, 2019, pp.
  1311--1318.

\bibitem{zhang2020temporal}
W.~Zhang and P.~Li, ``Temporal spike sequence learning via backpropagation for
  deep spiking neural networks,'' \emph{Advances in Neural Information
  Processing Systems}, vol.~33, pp. 12\,022--12\,033, 2020.

\bibitem{zheng2021going}
H.~Zheng, Y.~Wu, L.~Deng, Y.~Hu, and G.~Li, ``Going deeper with
  directly-trained larger spiking neural networks,'' in \emph{Proceedings of
  the AAAI Conference on Artificial Intelligence}, vol.~35, no.~12, 2021, pp.
  11\,062--11\,070.

\bibitem{deng2022temporal}
S.~Deng, Y.~Li, S.~Zhang, and S.~Gu, ``Temporal efficient training of spiking
  neural network via gradient re-weighting,'' \emph{arXiv preprint
  arXiv:2202.11946}, 2022.

\bibitem{garg2021dct}
I.~Garg, S.~S. Chowdhury, and K.~Roy, ``Dct-snn: Using dct to distribute
  spatial information over time for low-latency spiking neural networks,'' in
  \emph{Proceedings of the IEEE/CVF International Conference on Computer
  Vision}, 2021, pp. 4671--4680.

\bibitem{park2020t2fsnn}
S.~Park, S.~Kim, B.~Na, and S.~Yoon, ``T2fsnn: deep spiking neural networks
  with time-to-first-spike coding,'' in \emph{2020 57th ACM/IEEE Design
  Automation Conference (DAC)}.\hskip 1em plus 0.5em minus 0.4em\relax IEEE,
  2020, pp. 1--6.

\bibitem{kundu2021spike}
S.~Kundu, G.~Datta, M.~Pedram, and P.~A. Beerel, ``Spike-thrift: Towards
  energy-efficient deep spiking neural networks by limiting spiking activity
  via attention-guided compression,'' in \emph{Proceedings of the IEEE/CVF
  Winter Conference on Applications of Computer Vision}, 2021, pp. 3953--3962.

\bibitem{shen2022backpropagation}
G.~Shen, D.~Zhao, and Y.~Zeng, ``Backpropagation with biologically plausible
  spatiotemporal adjustment for training deep spiking neural networks,''
  \emph{Patterns}, vol.~3, no.~6, p. 100522, 2022.

\bibitem{rathi2020enabling}
N.~Rathi, G.~Srinivasan, P.~Panda, and K.~Roy, ``Enabling deep spiking neural
  networks with hybrid conversion and spike timing dependent backpropagation,''
  \emph{arXiv preprint arXiv:2005.01807}, 2020.

\bibitem{rathi2020diet}
N.~Rathi and K.~Roy, ``Diet-snn: Direct input encoding with leakage and
  threshold optimization in deep spiking neural networks,'' \emph{arXiv
  preprint arXiv:2008.03658}, 2020.

\bibitem{li2021free}
Y.~Li, S.~Deng, X.~Dong, R.~Gong, and S.~Gu, ``A free lunch from ann: Towards
  efficient, accurate spiking neural networks calibration,'' in
  \emph{International Conference on Machine Learning}.\hskip 1em plus 0.5em
  minus 0.4em\relax PMLR, 2021, pp. 6316--6325.

\bibitem{fang2021incorporating}
W.~Fang, Z.~Yu, Y.~Chen, T.~Masquelier, T.~Huang, and Y.~Tian, ``Incorporating
  learnable membrane time constant to enhance learning of spiking neural
  networks,'' in \emph{Proceedings of the IEEE/CVF International Conference on
  Computer Vision}, 2021, pp. 2661--2671.

\bibitem{han2020rmp}
B.~Han, G.~Srinivasan, and K.~Roy, ``Rmp-snn: Residual membrane potential
  neuron for enabling deeper high-accuracy and low-latency spiking neural
  network,'' in \emph{Proceedings of the IEEE/CVF conference on computer vision
  and pattern recognition}, 2020, pp. 13\,558--13\,567.

\bibitem{lu2020exploring}
S.~Lu and A.~Sengupta, ``Exploring the connection between binary and spiking
  neural networks,'' \emph{Frontiers in neuroscience}, vol.~14, p. 535, 2020.

\bibitem{deng2021optimal}
S.~Deng and S.~Gu, ``Optimal conversion of conventional artificial neural
  networks to spiking neural networks,'' \emph{arXiv preprint
  arXiv:2103.00476}, 2021.

\bibitem{werbos1988generalization}
P.~J. Werbos, ``Generalization of backpropagation with application to a
  recurrent gas market model,'' \emph{Neural networks}, vol.~1, no.~4, pp.
  339--356, 1988.

\bibitem{liu2001spike}
Y.-H. Liu and X.-J. Wang, ``Spike-frequency adaptation of a generalized leaky
  integrate-and-fire model neuron,'' \emph{Journal of computational
  neuroscience}, vol.~10, pp. 25--45, 2001.

\bibitem{kugele2020efficient}
A.~Kugele, T.~Pfeil, M.~Pfeiffer, and E.~Chicca, ``Efficient processing of
  spatio-temporal data streams with spiking neural networks,'' \emph{Frontiers
  in Neuroscience}, vol.~14, p. 439, 2020.

\bibitem{wu2021liaf}
Z.~Wu, H.~Zhang, Y.~Lin, G.~Li, M.~Wang, and Y.~Tang, ``Liaf-net: Leaky
  integrate and analog fire network for lightweight and efficient
  spatiotemporal information processing,'' \emph{IEEE Transactions on Neural
  Networks and Learning Systems}, vol.~33, no.~11, pp. 6249--6262, 2021.

\bibitem{xing2020new}
Y.~Xing, G.~Di~Caterina, and J.~Soraghan, ``A new spiking convolutional
  recurrent neural network (scrnn) with applications to event-based hand
  gesture recognition,'' \emph{Frontiers in neuroscience}, vol.~14, p. 590164,
  2020.

\bibitem{shrestha2018slayer}
S.~B. Shrestha and G.~Orchard, ``Slayer: Spike layer error reassignment in
  time,'' \emph{Advances in neural information processing systems}, vol.~31,
  2018.

\bibitem{krizhevsky2009learning}
A.~Krizhevsky, G.~Hinton \emph{et~al.}, ``Learning multiple layers of features
  from tiny images,'' 2009.

\bibitem{xu2015empirical}
B.~Xu, N.~Wang, T.~Chen, and M.~Li, ``Empirical evaluation of rectified
  activations in convolutional network,'' \emph{arXiv preprint
  arXiv:1505.00853}, 2015.

\bibitem{amir2017low}
A.~Amir, B.~Taba, D.~Berg, T.~Melano, J.~McKinstry, C.~Di~Nolfo, T.~Nayak,
  A.~Andreopoulos, G.~Garreau, M.~Mendoza \emph{et~al.}, ``A low power, fully
  event-based gesture recognition system,'' in \emph{Proceedings of the IEEE
  conference on computer vision and pattern recognition}, 2017, pp. 7243--7252.

\bibitem{li2017cifar10}
H.~Li, H.~Liu, X.~Ji, G.~Li, and L.~Shi, ``Cifar10-dvs: an event-stream dataset
  for object classification,'' \emph{Frontiers in neuroscience}, vol.~11, p.
  309, 2017.

\bibitem{plenz2021self}
D.~Plenz, T.~L. Ribeiro, S.~R. Miller, P.~A. Kells, A.~Vakili, and E.~L. Capek,
  ``Self-organized criticality in the brain,'' \emph{Frontiers in Physics},
  vol.~9, p. 639389, 2021.

\bibitem{cramer2020control}
B.~Cramer, D.~St{\"o}ckel, M.~Kreft, M.~Wibral, J.~Schemmel, K.~Meier, and
  V.~Priesemann, ``Control of criticality and computation in spiking
  neuromorphic networks with plasticity,'' \emph{Nature communications},
  vol.~11, no.~1, p. 2853, 2020.

\bibitem{beggs2003neuronal}
J.~M. Beggs and D.~Plenz, ``Neuronal avalanches in neocortical circuits,''
  \emph{Journal of neuroscience}, vol.~23, no.~35, pp. 11\,167--11\,177, 2003.

\bibitem{kim2023exploring}
Y.~Kim, Y.~Li, H.~Park, Y.~Venkatesha, A.~Hambitzer, and P.~Panda, ``Exploring
  temporal information dynamics in spiking neural networks,'' in
  \emph{Proceedings of the AAAI Conference on Artificial Intelligence},
  vol.~37, no.~7, 2023, pp. 8308--8316.

\end{thebibliography}

\section{Author Biography}
\begin{IEEEbiographynophoto}{Wenxuan Pan}received her B.Eng. degree from University of Electronic Science and Technology of China, Chengdu, Sichuan, China. She is currently a PhD student in the Brain-inspired Cognitive Intelligence Lab, Institute of Automation, Chinese Academy of Sciences, supervised by Prof. Yi Zeng. Her research interests focus on brain-inspired evolutionary models.
\end{IEEEbiographynophoto} 
\begin{IEEEbiographynophoto}{Feifei Zhao}received the BS degree in digital media technology from Northeastern University, in 2014, and the PhD degree in pattern recognition and intelligent systems from the University of Chinese Academy of Sciences, in 2019. She is currently an Associate Professor with the Brain-inspired Cognitive Intelligence Lab, Institute of Automation, Chinese Academy of Sciences. Her research interests include multi-brain areas coordinated learning and decision-making spiking neural network, brain-inspired developmental and evolutionay architecture search for SNNs. \end{IEEEbiographynophoto}
\begin{IEEEbiographynophoto}{Guobin Shen} received his B.Eng. degree from Sun Yat-sen University in Guangzhou, Guangdong, China. He is now a PhD candidate in the Brain-inspired Cognitive Intelligence Lab, at the Institute of Automation, Chinese Academy of Sciences, under the supervision of Prof. Yi Zeng. His research focuses on biologically-inspired learning algorithms and spiking neural network architecture design and training strategies. \end{IEEEbiographynophoto}
\begin{IEEEbiographynophoto}{Bing Han} received the B.Eng. degree in intelligent science and technology from University of Science and Technology Beijing, Beijing, China, in 2021. Now she is the Ph.D. candidate in the Brain-inspired Cognitive Intelligence Lab, Institute of Automation, Chinese Academy of Sciences, supervised by Prof. Yi Zeng. Her current research interests are brain-inspired structural development algorithms for spiking neural networks. \end{IEEEbiographynophoto}
\begin{IEEEbiographynophoto}{Yi Zeng}  
Yi Zeng obtained his Bachelor degree in 2004 and Ph.D degree in 2010 from Beijing University of Technology, China. He is currently a Professor and Director in the Brain-inspired Cognitive Intelligence Lab, Institute of Automation, Chinese Academy of Sciences (CASIA), China. He is a Principal Investigator in the Center for Excellence in Brain Science and Intelligence Technology, Chinese Academy of Sciences, China, and a Professor in the School of Future Technology, and School of Humanities, University of Chinese Academy of Sciences, China. His research interests include brain-inspired Artificial Intelligence, brain-inspired cognitive robotics, ethics and governance of Artificial Intelligence, etc.

 \end{IEEEbiographynophoto}
\end{document}